\documentclass[10pt,twocolumn,letterpaper]{article}

\usepackage[pagenumbers]{cvpr} 

\usepackage{graphics} 
\usepackage{epsfig} 
\usepackage{times} 
\usepackage{amsmath} 
\usepackage{amssymb}  
\usepackage{forest}
\usepackage{tikz}
\usepackage{xcolor}
\usepackage{multirow}
\usepackage{graphicx} 
\usepackage{amsmath}  
\usepackage{array}    
\usepackage{tabularx}              
\usepackage{caption}
\usetikzlibrary{shapes.geometric}
\usepackage{pifont}

\newcommand\extrafootertext[2]{%
    \bgroup
    \renewcommand\thefootnote{\fnsymbol{footnote}}%
    \renewcommand\thempfootnote{\fnsymbol{mpfootnote}}%
    \footnotetext[#1]{#2}%
    \egroup
}

\definecolor{darkgreen}{RGB}{0,127,0}
\definecolor{darkred}{RGB}{200,0,0}

\newcommand{\cmark}{\textcolor{darkgreen}{\ding{51}}}
\newcommand{\xmark}{\textcolor{darkred}{\ding{55}}}

\newcommand{\grupa}{\textcolor{darkgreen}{$\uparrow$}}

\newcommand{\datasetname}{\textsc{RoboSpatial}}

\definecolor{cvprblue}{rgb}{0.21,0.49,0.74}
\usepackage[pagebackref,breaklinks,colorlinks,allcolors=cvprblue]{hyperref}
\usepackage{graphicx}

\def\paperID{7547} 
\def\confName{CVPR}
\def\confYear{2025}

\title{\datasetname: Teaching Spatial Understanding to 2D and 3D Vision-Language Models for Robotics}

\author{Chan Hee Song$^{1*}$ \quad Valts Blukis$^2$ \quad Jonathan Tremblay$^2$ \quad Stephen Tyree$^2$ \quad Yu Su$^1$ \quad Stan Birchfield$^2$ \\\\ $^1$The Ohio State University, $^2$NVIDIA}

\begin{document}
\maketitle

\extrafootertext{1}{Corresponding author: \texttt{\small song.1855@osu.edu}. This work was partly done during the first author's internship at NVIDIA.}

\begin{abstract}

Spatial understanding is a crucial capability that enables robots to perceive their surroundings, reason about their environment, and interact with it meaningfully.
In modern robotics, these capabilities are increasingly provided by vision-language models. 
However, these models face significant challenges in spatial reasoning tasks, as their training data are based on general-purpose image datasets that often lack sophisticated spatial understanding.
For example, datasets frequently do not capture reference frame comprehension, yet effective spatial reasoning requires understanding whether to reason from ego-, world-, or object-centric perspectives.
To address this issue, we introduce \datasetname{}, a large-scale dataset for spatial understanding in robotics. 
It consists of real indoor and tabletop scenes, captured as 3D scans and egocentric images, and annotated with rich spatial information relevant to robotics.
The dataset includes 1M images, 5k 3D scans, and 3M annotated spatial relationships, and the pairing of 2D egocentric images with 3D scans makes it both 2D- and 3D- ready.
Our experiments show that models trained with \datasetname{} outperform baselines on downstream tasks such as spatial affordance prediction, spatial relationship prediction, and robot manipulation.

\end{abstract}
    
\section{Introduction}
\label{sec:intro}

\begin{figure}[tb] \centering
    \includegraphics[width=\linewidth]{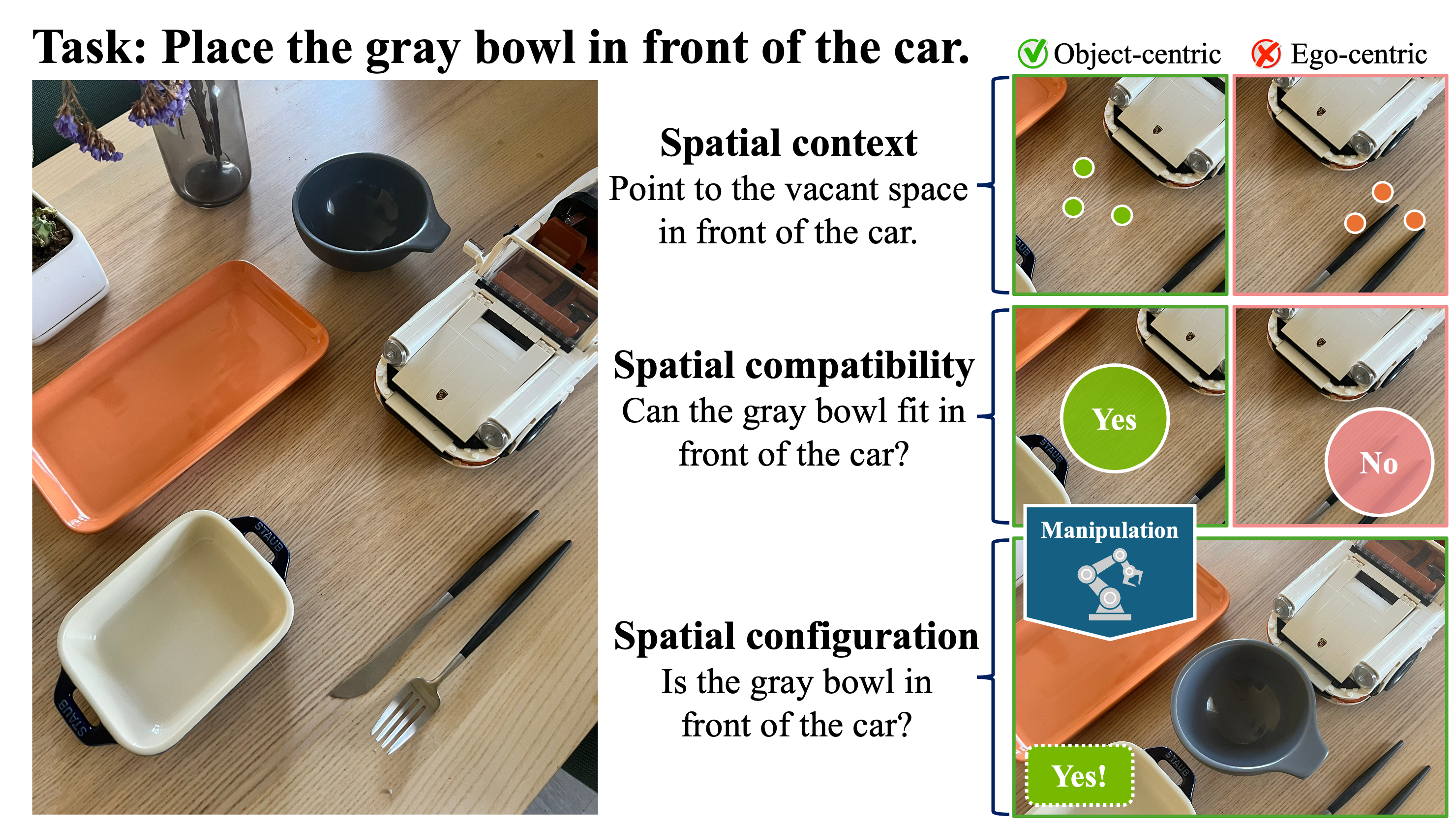}
    \caption{\datasetname{} dataset facilitates 3D spatial reasoning for robot manipulation. 
    This illustration demonstrates how a model trained on \datasetname{} enables human-aligned spatial reasoning within the correct reference frame, supporting task grounding, planning, and detection for manipulation tasks. 
    }
    \label{fig:intro}
    \vskip -10pt
\end{figure}

The rise of vision-language models (VLMs) has opened new opportunities for agents to interpret and act upon the visual world using natural language. 
VLMs have been adopted across a range of embodied settings, notably in robotics and augmented reality (AR).
In robotics, they have enabled grounded scene understanding~\cite{fangandliu2024moka, zook2024grsgeneratingroboticsimulation}, manipulation~\cite{open_x_embodiment_rt_x_2023}, and policy code generation~\cite{singh2023progprompt, codeaspolicies2022}, while in AR, they support tasks like object labeling~\cite{suglia-etal-2024-alanavlm}, action recognition~\cite{ego4d, mmgego4d}, and temporal grounding~\cite{huang2025egocentricvisionlanguagemodelbased}.

\begin{table*}[!t]
    \small
    \centering
    \renewcommand\arraystretch{0.8}
    \captionsetup{width=.85\textwidth} 

    \resizebox{\textwidth}{!}{
    \begin{tabular}{c|cccccccc} 
    \toprule
    Dataset & 3D scans & Embodied & Ref. frames & Compatibility & Domain & \#Scans & \#Images & \#Spatial QAs \\
    \midrule 
    EmbSpatial-Bench~\cite{du-etal-2024-embspatial} & \cmark & \cmark & \xmark & \xmark & Indoor & 277  & \,\,\,2k & \,\,\,4k \\
    Visual Spatial~\cite{liu-etal-2023-visualspatial} & \xmark & \xmark & \cmark & \xmark & MSCOCO & 0 & \,10k  & \,10k \\
    SpatialRGPT-Bench~\cite{cheng2024spatialrgpt} & \xmark & \xmark & \xmark & \cmark & Indoor, AV & 0 & 1.4k & 1.4k \\
    BLINK-Spatial~\cite{fu2024blink} & \xmark & \xmark & \cmark & \xmark & Generic & 0 & \,286 & \,286 \\
    What's up~\cite{kamath2023whatsup} & \xmark & \xmark & \xmark & \xmark & Generic & 0 & \,\,\,\,5k & \,10k \\    
    Spatial-MM~\cite{shiri-etal-2024-spatialmm} & \xmark & \xmark & \cmark & \xmark & Generic &  0 & 2.3k & 2.3k \\
    \midrule
    \datasetname& \cmark & \cmark & \cmark & \cmark & Indoor, tabletop & 5k & \,\,\,1M & \,\,\,3M \\
    \bottomrule 
    \end{tabular}
    }
    \captionof{table}{Comparison with other spatial reasoning datasets that include object-centric spatial relationships. 
    }
    \vskip -10pt

    \label{tab:dataset_comparison}
\end{table*}

VLMs can recognize objects, classify scenes, and even
provide general descriptions that capture high-level attributes.
However, despite significant recent advancements, VLMs~\cite{openai2022chatgpt, liu2023improvedllava, lin2023vila} 
still fall short in spatial understanding~\cite{yamada2024evaluatingspatialunderstandinglarge, kamath2023whatsup, shiri-etal-2024-spatialmm, OpenEQA2023, zhang2025do}.
They struggle with tasks that require interpreting nuanced spatial relationships between objects, such as describing where one object is in relation to another or determining the best location to place an item within a specific condition.
For example, while a model might accurately describe a ``bowl on the table," it lacks the ability to reason about where on the table the bowl is, where it should go to ensure accessibility or stability, or how it might fit among other objects. 
Furthermore, a critical limitation of existing VLM training datasets is their inability to capture \textit{reference frame} understanding (\textit{ref. frame}) --- the way we interpret spatial relationships changes drastically depending on whether we’re viewing from a first-person perspective, focusing on specific objects, or observing the entire scene, all of which are essential for real-world interactions.

These limitations highlight an ongoing challenge: bridging the gap between surface-level scene description and the deeper spatial comprehension necessary for intuitive interaction. 
Several recent efforts aim to address this by explicitly training VLMs on spatial reasoning tasks, yet many fall short of the demands posed by embodied or robotics settings.
For example, SpatialVLM~\cite{chen2024spatialvlm} and SpatialRGPT~\cite{cheng2024spatialrgpt} train VLMs to answer questions about distances and spatial relationships between objects, advancing spatial understanding at a conceptual level. 
However, these models are trained on datasets comprised of images from the internet, with annotations generated by perception models. 
As a result, they struggle to generalize to \textit{embodied} images—those captured by robot cameras within real-world environments, which often lack identifiable cues for an absolute scale.
Pointing models, such as RoboPoint \cite{yuan2024robopoint} and more recently Molmo~\cite{deitke2024molmopixmoopenweights}, take a different approach by training VLMs to produce grounded 2D coordinates that pinpoint object locations or free space within a scene.
However, these models lack understanding of real-world constraints, such as inferring object-centric reference frames for perspective-invariant reasoning, or accounting for the space required to place various objects. 
This results, for example, in failing to predict whether the gray bowl can fit in front of the car in Figure~\ref{fig:intro}.

This paper hypothesizes that a primary bottleneck limiting the effectiveness of VLMs in robotics is the scarcity of suitable training data, as highlighted by Table~\ref{tab:dataset_comparison}. 
To address this, we introduce \textbf{\datasetname}, a dataset designed specifically to facilitate spatial understanding in VLMs for robotic applications. 
The proposed approach leverages annotated indoor scene and tabletop RGBD datasets, transforming them into targeted question-answer pairs designed to probe spatial reasoning skills critical for robotics.

We categorize the questions into three types, each serving a distinct purpose. 
\textbf{Spatial context} focuses on identifying empty space or support surfaces in the environment that can accommodate other objects. 
These questions are formulated as point-prediction tasks, challenging the model to determine appropriate locations within free space where an object can be placed—for example, ``Where on the table can I put the plate?'' 
\textbf{Spatial compatibility} builds on the identified empty space to assess whether a given area can feasibly support the placement of a specific object, ensuring sufficient size and fit. 
These questions are posed in a binary format, such as ``Can the chair be placed in front of the table?'' \textbf{Spatial configuration} examines the relative spatial relationships between two objects. 
These questions use a binary format to determine whether a spatial relation holds, such as ``Is the mug to the left of the laptop?''

To enhance the model’s ability to interpret spatial instructions from different perspectives, each question-answer pair in \datasetname{} is posed from three distinct reference perspectives/frames: 
(a)~\textbf{Ego-centric} from the observer’s perspective at the camera pose, 
(b)~\textbf{World-centric} grounded in a global world frame, and
(c)~\textbf{Object-centric} based on a reference frame attached to the focal object.
This multi-frame approach enables models to handle complex spatial instructions more flexibly, preparing them to better generalize to dynamic robotic contexts.
Applying our methodology to existing indoor scene and tabletop datasets, we generate both a comprehensive training dataset and a benchmark for spatial question answering in robotics.
\datasetname{} contains around \textbf{1M} images, \textbf{5k} 3D scans, and \textbf{3M} annotated spatial relationships, with paired 2D egocentric images and 3D scans to make it both 2D- and 3D- ready. 

To validate the effectiveness of \datasetname{}, comprehensive experiments were conducted using multiple state-of-the-art (SOTA) 2D and 3D VLMs.
Results demonstrate that models trained on \datasetname{} exhibit significantly improved spatial reasoning capabilities, consistently outperforming baseline methods on the evaluation benchmark \textbf{\datasetname{}-Val}, a held-out validation subset derived from the heuristically generated \datasetname{} dataset.
To further assess the generalization and robustness of these trained VLMs, additional evaluations were conducted using three complementary benchmarks: \textbf{\datasetname{}-Home}, a manually collected dataset consisting of paired RGB and depth images, and two external benchmarks, BLINK-Spatial~\cite{fu2024blink} and SpatialBot~\cite{cai2024spatialbot}. 
These benchmarks rigorously test spatial reasoning skills in practical robotic tasks, including object rearrangement and contextual question answering in indoor environments, while also examining the models' capacity to generalize to novel spatial reasoning scenarios beyond the original training data. 
Across all benchmarks, models trained on \datasetname{} consistently outperformed baseline methods, demonstrating the broad utility of the dataset.
Leveraging the 3D-ready design of \datasetname{}, direct comparisons between the spatial reasoning performance of 2D and 3D VLMs were also performed. 
Although initial results indicate potential advantages for 3D models, differences in pretraining data and base LLM architectures among models render the comparison inconclusive. 
\datasetname{} is specifically designed to support both 2D and 3D research directions, enabling future studies to address these differences more conclusively.

\noindent \textbf{Our contributions} are threefold:
\begin{itemize}[noitemsep, topsep=0pt, parsep=0pt] 
    \item A new training dataset, \datasetname{}, comprising images and 3D scans paired with spatial questions and answers, accompanied by an evaluation benchmark, \datasetname{}-Val, a held-out validation set. Additionally, we introduce \datasetname{}-Home, a manually collected and annotated dataset designed specifically for assessing real-world spatial reasoning in indoor environments. These datasets uniquely incorporate multiple reference frames, object-object spatial relationships, object-space relationships, and object compatibility. We make the data and code for generating the dataset from 3D annotated scenes publicly available\footnote{\url{https://chanh.ee/RoboSpatial/}}.
    \item VLMs trained on \datasetname{} demonstrate superior spatial reasoning, outperforming SOTA baselines on language-guided robot manipulation and indoor scene question answering.
    \item Comprehensive experiments assessing spatial reasoning capabilities in both 2D and 3D VLMs, comparing the difference between SOTA VLMs in real-world spatial tasks. 
\end{itemize}

\section{Related Work}
\label{sec:related_works}

\noindent\textbf{VLMs for Robotics.} Vision-language models (VLMs) have emerged as pivotal tools in robotics, enabling systems to interpret and act upon complex visual and textual information. 
By integrating visual perception with language understanding, VLMs facilitate more intuitive human-robot interactions and enhance autonomous decision-making capabilities.
Recent advancements have demonstrated the potential of VLMs in various robotic applications. 
For instance, vision-language-action models (VLAs)~\cite{kim24openvla, rt22023arxiv, octo_2023} enable robots to interpret and execute complex instructions and output executable robot actions.
Additionally, VLMs like GPT-4v~\cite{openai2022chatgpt} have been utilized for high-level task planning~\cite{wake2023gpt}, allowing robots to generate detailed action sequences from natural language instructions. 
Furthermore, VLMs have been used for keypoint/mask prediction~\cite{huang2024rekep, wi2023calamari, google2024pivot}, error analysis~\cite{duan2024ahavisionlanguagemodeldetectingreasoning, Song_2022_CVPR}, grasp pose prediction~\cite{huang2024copageneralroboticmanipulation}.
Despite these advancements, integrating VLMs~\cite{cai2024spatialbot, cheng2024spatialrgpt, yuan2024robopoint} into robotic systems presents challenges. 
One significant hurdle is the need for precise spatial reasoning to navigate and manipulate objects effectively.
While VLMs excel in understanding and generating language, their ability to comprehend and reason about spatial relationships in dynamic environments remains limited~\cite{yamada2024evaluatingspatialunderstandinglarge, xu2024evaluatinglargelanguagemodels, wang2024pictureworththousandwords}.
Therefore, \datasetname{} aims to tackle this gap by presenting a large scale pretraining and evaluation setup for teaching spatial understanding to VLM for robotics.

\begin{figure}[!t] 
    
    \centering
    \includegraphics[width=1\linewidth]{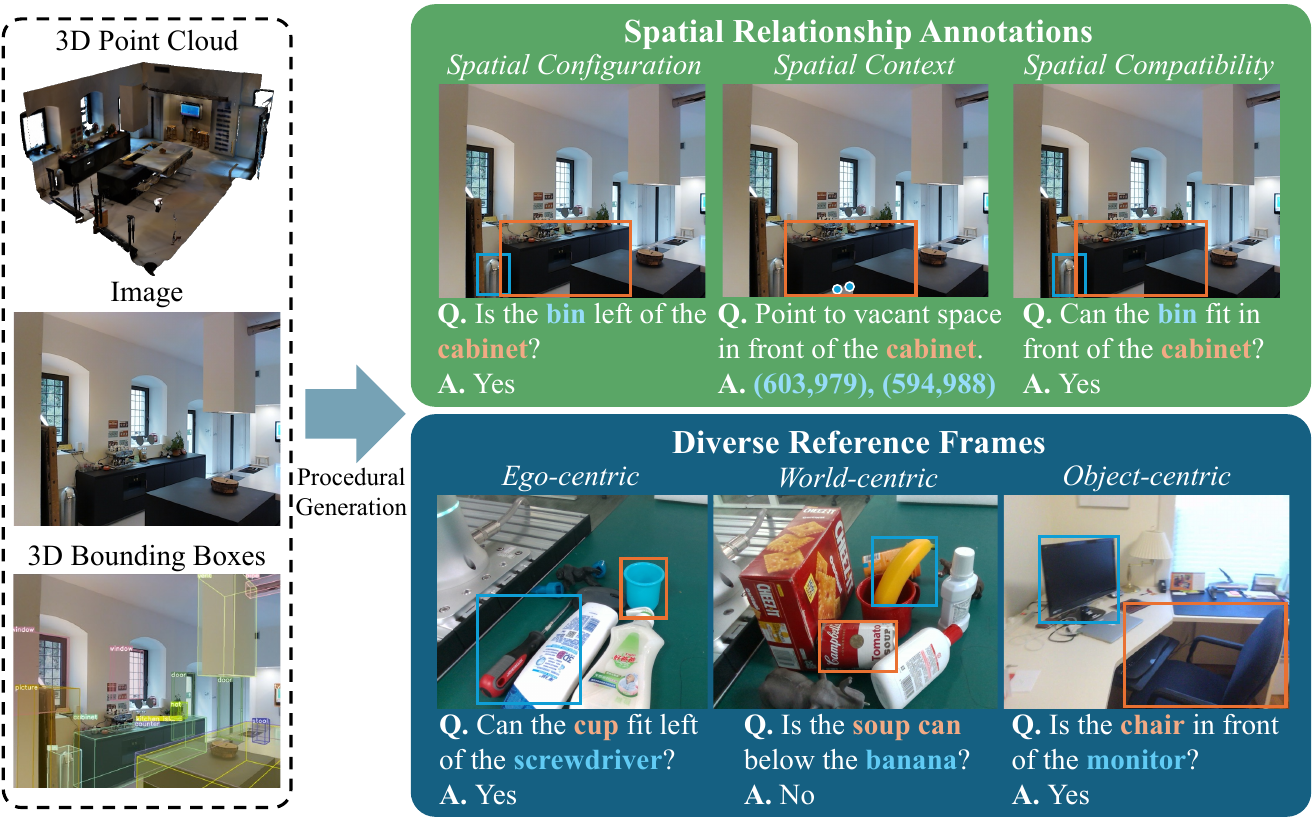}
    \caption{Overview of the \datasetname{} dataset. We automatically generate spatial relationship annotations from existing datasets with 3D point clouds, egocentric images, and 3D bounding box annotations. We create question/answer pairs covering three classes of spatial relationships, three spatial reference frames, and both binary (yes/no) and numeric (\eg 2D image points) answers.
    From 1M images and 5k scans, we generate over 3M spatial question/answer pairs.}
    \label{fig:dataset_intro}
    \vskip -10pt
\end{figure}

\noindent\textbf{Spatial Understanding with VLMs.} Spatial understanding has been implicitly and explicitly part of various vision and question answering tasks~\cite{fu2024blink, scanqa, jia2024sceneversescaling3dvisionlanguage, suhr2019nlvr2visualbiasanalysis, salewski2022clevrx, krishna2017visualgenome, johnson2017clevr, hudson2019gqa}. While many benchmarks and methods have been proposed, they often come with limitations: some focus exclusively on simulations~\cite{szymanska2024space3dbench} or generic images~\cite{liu-etal-2023-visualspatial, rajabi2024groundedvisualspatialreasoning, cheng2024spatialrgpt, chen2024spatialvlm, fu2024blink, kamath2023whatsup, shiri-etal-2024-spatialmm, ranasinghe_spatial}, others are difficult to evaluate due to their reliance on free-form text outputs~\cite{szymanska2024space3dbench, du-etal-2024-embspatial, linghu2024multi-modalsituated}, some rely on complete 3D scans~\cite{zhang2024spartun3dsituatedspatialunderstanding, man2024situation3d, ma2022sqa3d, linghu2024multi-modalsituated}, and others do not account for reference frames~\cite{zhang2024spartun3dsituatedspatialunderstanding, man2024situation3d, ma2022sqa3d, linghu2024multi-modalsituated, chen2024spatialvlm, cheng2024spatialrgpt, fu2024blink, ranasinghe_spatial}. 
Furthermore, many fail to address actionable, robotics-relevant spatial relationships such as spatial compatibility and context~\cite{du-etal-2024-embspatial, fu2024blink, wang2023embodiedscan, shiri-etal-2024-spatialmm, kamath2023whatsup, linghu2024multi-modalsituated, ranasinghe_spatial}.

Inspired by prior works on spatial reasoning~\cite{liu-etal-2023-visualspatial, kamath2023whatsup}—where the impact of reference frames and spatial configurations was explored in generic images~\cite{lin2015microsoftcococommonobjects, hudson2019gqa}—we extend spatial understanding to a robotics-specific context with actionable spatial relationships such as spatial compatibility and spatial context. 
Our aim is to enable direct application to robotic workflows, such as task planning and verification.

To achieve this, we have developed a large-scale 2D/3D ready training dataset using our automated data generation pipeline.
We further show how \datasetname{} can be used to teach spatial reasoning to a suite of vision-language models (VLMs) in in-domain and out-of-domain spatial reasoning datasets.
We hope these resources lower the barrier to entry for exploring spatial understanding tailored to robotics.

\section{Approach}

We begin by explaining the selection of three spatial relationships:  spatial context, spatial compatibility, and spatial configuration.
Next, we describe the data generation pipeline used to construct the \datasetname{}. 
Figure~\ref{fig:dataset_intro} provides an overview of the dataset.
%

\subsection{Spatial Relationships}
The dataset is organized around three core spatial relationships that we believe address the essential aspects of spatial reasoning for robotic tasks: spatial \textit{context}, spatial \textit{compatibility}, and spatial \textit{configuration}.
\textit{Context} allows robots to assess the relationship between objects and their surrounding space, facilitating the identification of empty or occupied areas, which is relevant for downstream applications such as path planning and obstacle avoidance.
\textit{Compatibility} focuses on whether objects can coexist or interact without conflict in a given space, which is vital for object placement, assembly, and operational safety.
\textit{Configuration} enables robots to understand and interpret the relative positioning of objects, which is crucial for directing navigation, manipulation, and interaction within complex environments. 
Together, these spatial relationships provide a more nuanced and practical framework for robotic applications than metrics like distance---which is hard to normalize across different scales, environments, and tasks---thereby enabling robots to perform complex tasks with greater reliability.

\subsection{Dataset Generation}
\label{sec:data_gen}

The goal of the data construction pipeline is to generate a large-scale, high-accuracy spatial relationship dataset with minimal human intervention, using automatic heuristics grounded in 3D geometry and 2D image views.

The pipeline takes as input a scene dataset $\mathcal{D}_{s}$ that contains RGB images, camera poses (both extrinsic and intrinsic parameters), and oriented 3D bounding box annotations with semantic object labels. 
The output is a spatial reasoning dataset $\mathcal{D}$, where each entry $d_i = \langle I_i, q_i, a_i, l_i \rangle$ consists of an image $I_i$, a question $q_i$, an answer $a_i$, and a reference frame label $l_i \in \{\textit{ego}, \textit{world}, \textit{object}\}$. Each question is derived from one of three spatial reasoning categories: spatial configuration, spatial context, or spatial compatibility. 
To support reliable object reference resolution, we also generate an auxiliary object grounding dataset that links object descriptions to 2D bounding boxes.

To improve clarity and reproducibility, we describe the data generation pipeline in two main stages. We also separate the reasoning logic used for extracting 3D relationships from that used for generating 2D image-space targets.

\subsubsection{Stage 1: 3D Spatial Relation Extraction}
\label{subsec:spatial_relationship_def}

The first stage involves extracting spatial relationships between objects or between objects and free space, based on 3D geometry. Each spatial relation is defined as $s_i = \langle I_i, a_i, t_i, r_i, l_i \rangle$, where $I_i$ is the source image, $a_i$ is the anchor object, $t_i$ is the target object or a sampled point in free space, $r_i \in \{\textit{left}, \textit{right}, \textit{above}, \textit{below}, \textit{front}, \textit{behind}\}$ is the relation preposition, and $l_i \in \{\textit{ego}, \textit{world}, \textit{object}\}$ denotes the reference frame.

We use oriented 3D bounding boxes, provided by the source dataset, to compute spatial relationships. Each bounding box includes both the 3D location and heading of the object. 
The object’s orientation is defined by the heading vector of the bounding box, aligned with the object’s front-facing direction. 
Using this orientation, we determine the appropriate directional region (\eg, front, left) relative to the reference frame. 
For instance, a relation such as ``in front of (anchor object) (object frame)'' refers to the positive direction along the anchor object’s heading vector. 
These relationships are calculated independently for each of the three reference frames: the world frame is aligned with the dataset-level coordinate system; the ego frame is defined by the camera pose (\ie, camera-centered); and the object frame is defined by the local orientation of the anchor object.

The camera extrinsics are used to transform coordinates between reference frames. 
Although the method does not require point clouds or meshes, it relies on camera intrinsics and extrinsics to project between 2D and 3D and to ensure consistent reference frame reasoning. 
For each spatial configuration task, we evaluate all visible object pairs that appear uniquely in the image, avoiding duplicate instances to minimize ambiguity. 
The resulting relationships are binary (True/False) and specify whether the spatial condition holds for the given object pair.

\subsubsection{Stage 2: 2D Spatial Point and Region Sampling}

In the second stage, we generate 2D image-space annotations for spatial context and spatial compatibility tasks. 
These rely on the 3D bounding box layout and calibrated camera parameters to map spatial relationships into image coordinates.

For spatial context, we construct a top-down occupancy map of the scene by marking regions occupied by 3D bounding boxes. 
We then randomly sample 3D points in empty space that lie in a specified directional relation to the anchor object, following the same frame-dependent heuristics as in the configuration task. 
These points are projected into the image plane using the camera intrinsics. To ensure the points are valid, we filter out samples that are obstructed or occluded based on line-of-sight from the camera. 
Specifically, we perform raycasting from the camera center to each sampled 3D point, and discard points whose rays intersect any occupied bounding box volumes before reaching the target location. 
The final answer is a list of 2D $(x, y)$ image coordinates that satisfy the spatial context constraint.

Spatial compatibility extends this idea by checking whether a target object can fit within the sampled region. 
We simulate placing a virtual bounding box, matching the size of the target object, at the candidate location on the ground plane. 
A region is considered compatible if the simulated placement does not intersect with any existing bounding boxes in the scene and provides at least a 10 cm margin along each axis. 
The simulation allows for translation and in-plane rotation of the object. 
The answer to this task is binary (True/False), indicating whether the region can accommodate the object.

\subsubsection{Question-Answer Generation}

Once the spatial relations $\{s_i\}$ have been extracted, we generate corresponding question-answer pairs $\{d_i\}$ using structured templates. Each question follows the format:
$\textsc{\{target\} \{relation\} \{anchor\} \{ref. frame\}}$
%
where the relation and frame are defined in Section~\ref{subsec:spatial_relationship_def}. To ensure that models learn from visual grounding rather than linguistic priors, we use deterministic templates that avoid ambiguity and minimize reliance on commonsense.

Each spatial relation type—context, compatibility, and configuration—has a corresponding question format. 
Configuration and compatibility tasks result in binary (True/False) answers.
Context questions produce a list of valid 2D coordinates in image space.

Correctly resolving which object is being referred to in a spatial question is essential for reliable spatial understanding. To reduce errors arising from incorrect object identification, we additionally generate an auxiliary object grounding dataset that links object descriptions to 2D bounding boxes in the image. These grounding annotations are derived by projecting 3D bounding boxes into image space using camera intrinsics and extrinsics. This supervision helps models more accurately resolve references during spatial reasoning and is included during training. See Appendix~\ref{appendix:object_grounding} for details.

Using this pipeline, we generate around 3 million spatial relationships and their associated question-answer pairs. This scale is an order of magnitude larger than prior spatial reasoning datasets (see Table~\ref{tab:dataset_comparison}). 

\section{Experiments}

\subsection{Setup}
\label{sec:implementation}

We apply the data generation pipeline to three scene datasets—ScanNet~\cite{dai2017scannet}, Matterport3D~\cite{chang2017matterport3d}, and 3RScan~\cite{Wald2019RIO}—and two tabletop datasets—HOPE~\cite{tyree2022hope} and GraspNet-1B~\cite{fang2020graspnet}. 
We retrieve 3D bounding box annotations and embodied images from EmbodiedScan~\cite{wang2023embodiedscan}, and generate a large-scale spatial reasoning dataset covering diverse indoor environments: larger scenes for navigation and smaller object-centric setups for manipulation.

In total, \datasetname{} includes approximately 3M spatial QA pairs across 5k 3D scans and 1M images.
(Table~\ref{tab:datasets} provides a breakdown; values are rounded to the nearest thousand for clarity.) 

\begin{table}[t]
\centering
\resizebox{\linewidth}{!}{
\begin{tabular}{lcccc}
\toprule
\textbf{Dataset}          & \textbf{Type}    & \textbf{Splits} & \textbf{Images} & \textbf{QA pairs}\\ 
\midrule
\multirow{2}{*}{Indoor}     & Train            & 4916 scans      & 883k images &    2.7M       \\ 
                          & Validation             & 40 scans       & 1k images       &    3k  \\ 
\midrule
\multirow{2}{*}{Tabletop}     & Train            & 190 scenes       & 76k images   &     220k     \\ 
                          & Validation             & 77 scenes       & 355 images        &   3k   \\ 
\bottomrule
\end{tabular}
}
\caption{Dataset splits for indoor and tabletop dataset. Detailed data statistics are in the Appendix.}

\label{tab:datasets}
\vskip -10pt
\end{table}



\subsubsection{Trained 2D/3D VLMs}

\noindent \textbf{2D VLMs.} \ We evaluate several vision-language models (VLMs) using RGB-only image inputs.
Our selected base VLMs are VILA-1.5-8B~\cite{lin2023vila} and LLaVA-NeXT-8B~\cite{liu2023improvedllava}. 
We also include three specialized models: 
SpaceLLaVA-13B (a community version of SpatialVLM~\cite{chen2024spatialvlm}), 
RoboPoint-13B~\cite{yuan2024robopoint} (trained to predict points in empty space given an object reference), 
and Molmo-7B~\cite{deitke2024molmopixmoopenweights} (designed for pointing and counting). 
We also include GPT-4o~\cite{openai2022chatgpt} as a closed-source baseline. 
We omit models such as SpatialRGPT~\cite{cheng2024spatialrgpt} that depend on external mask inputs, as they bypass the object grounding challenge.

\noindent \textbf{3D VLMs.} \ Models operating over 3D data must handle richer, more complex spatial representations. 
We include two models that process 3D inputs: 
3D-LLM~\cite{3dllm}, which reconstructs colored 3D point clouds from multi-view RGB images, 
and LEO~\cite{huang2023embodied}, which operates on segmented colored point clouds of individual objects.
These models allow us to explore spatial reasoning when models consume RGBD or point cloud representations directly.

\noindent \textbf{Fine-tuning.} \ We evaluate models in both zero-shot and fine-tuned settings, using \datasetname{} to fine-tune open-source models.
To mitigate failure cases arising from poor object grounding, we also include an auxiliary grounding dataset during training, which provides additional supervision for object reference resolution.
This auxiliary dataset does not contribute to spatial reasoning performance. (See Appendix for ablation experiments.)

\begin{table*}[t]
    \centering
    
    \tabcolsep 5pt
    \renewcommand\arraystretch{1}
    \captionsetup{width=0.9\textwidth} 
    \resizebox{0.9\textwidth}{!}{
    \begin{tabular}{lccccccccc}
    \toprule
    \multirow{2}{*}{\textbf{Model}} &  \multicolumn{3}{c}{\textbf{Indoor}} & \multicolumn{3}{c}{\textbf{Tabletop}} & \multicolumn{3}{c}{\textbf{Average}}\\
     \cmidrule(r){2-4} \cmidrule(r){5-7} \cmidrule(r){8-10} &  \textbf{Configuration}  & \textbf{Context} & \textbf{Compatibility} & \textbf{Configuration}  & \textbf{Context} & \textbf{Compatibility}  & \textbf{Indoor}  & \textbf{Tabletop}  &\textbf{Total} \\
    \midrule 
    \multicolumn{10}{c}{\textit{Open-source VLMs}} \\
    \multicolumn{10}{c}{\textbf{2D VLMs}} \\
    VILA~\cite{lin2023vila} & 54.7\,\,\,\, & 18.3\,\,\,\,  & 56.3\,\,\,\, &  45.1\,\,\,\, & 13.2\,\,\,\, & 53.8\,\,\,\, &  43.1\,\,\,\, & 37.4\,\,\,\, &  40.2\,\,\,\,   \\
    \,\,\,+\datasetname{} &  71.4~\grupa & 45.9~\grupa &  77.2~\grupa & 71.8~\grupa & 43.7~\grupa &  73.3~\grupa  & 64.8~\grupa & 62.9~\grupa & 63.9~\grupa \\
    \cmidrule(r){1-1}
    LLaVA-NeXT~\cite{liu2023improvedllava} & 48.9\,\,\,\,  & 12.5\,\,\,\, &  32.7\,\,\,\, & 48.3\,\,\,\, & 8.4\, & 30.9\,\,\,\,  & 31.4\,\,\,\,  & 29.2\,\,\,\, &   30.3\,\,\,\,  \\
    \,\,\,+\datasetname{} &  69.3~\grupa & 41.3~\grupa &  70.5~\grupa & 70.7~\grupa & 44.8~\grupa &  66.1~\grupa  & 60.4~\grupa & 60.5~\grupa & 60.5~\grupa  \\
    \cmidrule(r){1-1}
    SpaceLLaVA~\cite{chen2024spatialvlm} & 52.6\,\,\,\,  & 15.3\,\,\,\, &  49.0\,\,\,\, & 66.5\,\,\,\, & 12.2\,\,\,\, & 60.1\,\,\,\,  &  38.9\,\,\,\, & 46.2\,\,\,\, & 43.6\,\,\,\,  \\
    \,\,\,+\datasetname{} &  76.0~\grupa & 50.7~\grupa &  76.6~\grupa & 74.9~\grupa & 46.4~\grupa &   70.5~\grupa  &  67.8~\grupa & 63.6~\grupa & 65.7~\grupa \\
    \cmidrule(r){1-1}
    RoboPoint~\cite{yuan2024robopoint} &  39.0\,\,\,\, & 41.4\,\,\,\, & 38.3\,\,\,\,  &  37.9\,\,\,\, &  31.6\,\,\,\, & 45.2\,\,\,\, &  39.6\,\,\,\, & 38.2\,\,\,\, & 38.9\,\,\,\, \\
    \,\,\,+\datasetname{} &  72.2~\grupa & \textbf{68.9}~\grupa &  72.1~\grupa &  70.3~\grupa & \textbf{61.7}~\grupa &  78.4~\grupa & 71.0~\grupa & 70.1~\grupa & 70.6~\grupa  \\ \\

    \multicolumn{10}{c}{\textbf{3D VLMs}} \\
    3D-LLM~\cite{3dllm} & 54.5\,\,\,\, & 8.1\, & 53.6\,\,\,\,  &  59.2\,\,\,\, &  10.6\,\,\,\, &  57.4\,\,\,\, &  37.6\,\,\,\, & 42.4\,\,\,\, &  40.0\,\,\,\, \\ 
        \,\,\,+\datasetname{} &  76.3~\grupa &  35.4~\grupa &  77.5~\grupa & 76.2~\grupa &  46.8~\grupa &  75.0~\grupa &  63.1~\grupa & 66.0~\grupa & 64.6~\grupa \\ 
    \cmidrule(r){1-1}
    LEO~\cite{huang2023embodied} &  56.1\,\,\,\,  & 11.3\,\,\,\, &  58.3\,\,\,\,  &  60.8\,\,\,\, &  11.1\,\,\,\, &   59.3\,\,\,\, &  41.9\,\,\,\, &  43.7\,\,\,\,  & 42.8\,\,\,\, \\
    \,\,\,+\datasetname{} &  \textbf{80.2}~\grupa &  56.7~\grupa &  \textbf{82.5}~\grupa & \textbf{78.1}~\grupa &  55.2~\grupa &  \textbf{78.9}~\grupa & \textbf{73.1}~\grupa  & \textbf{70.7}~\grupa &  \textbf{71.9}~\grupa \\ 
    
    \midrule
    \multicolumn{10}{c}{\textit{Not available for fine-tuning}} \\
    \multicolumn{10}{c}{\textbf{2D VLMs}} \\
    Molmo~\cite{deitke2024molmopixmoopenweights} & 40.6\,\,\,\, & 48.2\,\,\,\, & 60.0\,\,\,\, & 61.5\,\,\,\, & 35.8\,\,\,\, & 54.6\,\,\,\, & 49.6\,\,\,\,  & 50.6\,\,\,\, & 50.1\,\,\,\, \\
    GPT-4o~\cite{openai2022chatgpt} & 63.5\,\,\,\,  & 25.1\,\,\,\, & 59.4\,\,\,\,  &  62.3\,\,\,\, & 27.9\,\,\,\, & 66.8\,\,\,\, & 49.3\,\,\,\,  & 52.3\,\,\,\, & 50.8\,\,\,\, \\ 
    \bottomrule
    \end{tabular}
    }
    \caption{Results of existing 2D/3D VLMs on a held-out validation split (\datasetname-Val) of images and scans.  All methods, for all tasks, perform better (\grupa) when fine-tuned on \datasetname{}. The best result for each column is bolded.}

    \label{tab:main_results}
    
\end{table*}

\begin{table}[t]
    \centering
    
    \tabcolsep 2pt
    \renewcommand\arraystretch{1}

    \resizebox{\linewidth}{!}{
    \begin{tabular}{lccccccc}
    \toprule
    \multirow{2}{*}{\textbf{Model}} &  \multicolumn{3}{c}{\textbf{\datasetname-Home}} & \textbf{BLINK} & \textbf{SpatialBench} \\
     \cmidrule(r){2-4} \cmidrule(r){5-5} \cmidrule(r){6-6} &  \textbf{Configuration}  & \textbf{Context} & \textbf{Compatibility} & \textbf{Accuracy} & \textbf{Accuracy} \\
    \midrule 
    \multicolumn{6}{c}{\textbf{2D VLMs}} \\
    VILA~\cite{lin2023vila} &  57.8\,\,\,\, & 0.0\,  & 69.0\,\,\,\, &   72.7\,\,\,\, &   53.0\,\,\,\, \\
    \,\,\,+\datasetname & 65.9~\grupa  &  15.6~\grupa & 78.0~\grupa &  79.7~\grupa & \textbf{73.6}~\grupa \\
    \cmidrule(r){1-1}
    LLaVA-NeXT~\cite{liu2023improvedllava} & 68.3\,\,\,\,  & 0.0\,  & 70.5\,\,\,\, &  71.3\,\,\,\,  & 55.9\,\,\,\, \\
    \,\,\,+\datasetname & \textbf{78.9}~\grupa  & 19.7~\grupa  & 80.1~\grupa &  79.0~\grupa & 70.6~\grupa \\
    \cmidrule(r){1-1}
    SpaceLLaVA~\cite{chen2024spatialvlm} & 61.0\,\,\,\,  & 2.5\,  & 41.0\,\,\,\, & 76.2\,\,\,\, & 47.1\,\,\,\,  \\
    \,\,\,+\datasetname & 71.6~\grupa  & 13.1~\grupa  & 72.4~\grupa &   \textbf{81.8}~\grupa & 67.7~\grupa \\
    \cmidrule(r){1-1}
    RoboPoint~\cite{yuan2024robopoint} & 69.9\,\,\,\,  &  19.7\,\,\,\, & 70.5\,\,\,\, &  63.6\,\,\,\, &  44.1\,\,\,\,  \\
    \,\,\,+\datasetname & 78.0~\grupa  & \textbf{31.1}~\grupa  & \textbf{81.0}~\grupa &   70.6~\grupa &   64.7~\grupa \\
    
    \multicolumn{6}{c}{\textbf{3D VLMs}} \\
    3D-LLM~\cite{3dllm} &  39.8\,\,\,\, &  0.0\, & 35.2\,\,\,\,  &  N/A\,\,\,\, &  N/A\,\,\,\,  \\
    \,\,\,+\datasetname & 55.2~\grupa  & 8.2~\grupa  & 52.3~\grupa &  N/A\,\,\,\, &  N/A\,\,\,\, \\
    \cmidrule(r){1-1}
    LEO~\cite{huang2023embodied} &  51.2\,\,\,\, &  0.0\, & 38.1\,\,\,\,  &  N/A\,\,\,\, &  N/A\,\,\,\,\\
    \,\,\,+\datasetname &  64.2~\grupa &  10.0~\grupa & 57.1~\grupa &   N/A\,\,\,\,&  N/A\,\,\,\, \\
    
    \midrule
    \multicolumn{6}{c}{\textit{Not available for fine-tuning}} \\
    Molmo~\cite{deitke2024molmopixmoopenweights} &  58.6\,\,\,\, &  0.1\, & 18.1\,\,\,\, & 67.1\,\,\,\, & 55.9\,\,\,\,   \\
    GPT-4o~\cite{openai2022chatgpt}   & 77.2\,\,\,\,  & 5.7\,  & 58.1\,\,\,\, &  76.2\,\,\,\,  &  70.6\,\,\,\,  \\ 
    \bottomrule
    \end{tabular}
    }
    \caption{Results on an out-of-domain test split comparing prior art VLMs. The results show improved (\grupa) spatial understanding capabilities on similar domains. Bolded number is the best result for the column.
    }
    \label{tab:downstream_results}
    
    \vskip -10pt
\end{table}

\subsubsection{Spatial Understanding Evaluation}

We evaluate spatial reasoning capabilities using \datasetname-Val, a held-out validation subset of \datasetname{} sampled from scans that are entirely unseen during training. This benchmark comprises 6,000 heuristically generated questions from our data generation pipeline, with 2,000 questions per spatial relation type.
Questions fall into two categories: binary yes/no questions and coordinate prediction tasks. For yes/no questions, we report accuracy. For coordinate predictions, we evaluate whether the model’s predicted 3D location lies within the convex hull of a reference point set derived from scene geometry.

While this convex hull criterion provides a well-defined geometric target, it is arguably overly strict—\eg, predictions near but just outside the boundary are marked incorrect. As a result, reported scores represent a conservative estimate of each model’s spatial understanding.
\datasetname{}-Val serves as the primary benchmark for comparing 2D and 3D VLMs trained on \datasetname{}, enabling controlled evaluation within the same data distribution. Results are presented in Table~\ref{tab:main_results}.

\subsubsection{Cross-Dataset Generalization Evaluation}

To assess generalization across environment types, we partition the training data into indoor-scene and tabletop subsets.
Models are trained on one type and evaluated on held-out datasets from the other.
Despite differing object distributions and scene layouts, we observe a positive synergy between indoor and tabletop environments: training on one environment type improves spatial reasoning on the other, as shown in Table~\ref{tab:dataset_generalization}.

\begin{figure*}[tb]
    \centering
    \includegraphics[width=0.9\textwidth]{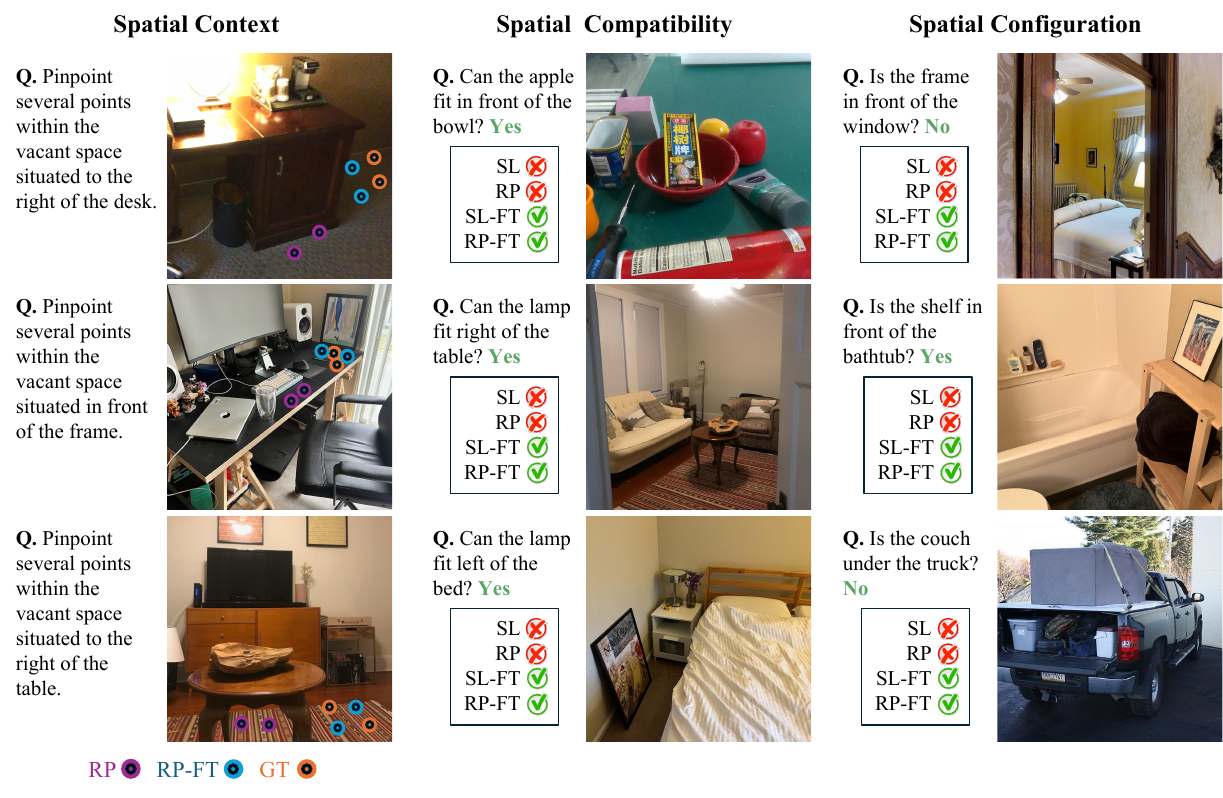}
    \caption{In-domain (\datasetname-Val, top) and out-of-domain (\datasetname-Home, BLINK~\cite{fu2024blink}, middle and bottom) results for \datasetname-trained models. Two models shown: SL (SpaceLLaVA~\cite{chen2024spatialvlm}) and RP (RoboPoint~\cite{yuan2024robopoint}); the -FT suffix indicates fine-tuning on \datasetname. Correct answers in green. All images except bottom-right in the out-of-domain rows are from \datasetname-Home.}
    \label{fig:exp}
    \vskip -10pt
\end{figure*}

\subsubsection{Out-of-Domain Evaluation}

To test the out-of-domain transferability of \datasetname{}-trained models, we evaluate on three benchmarks: \datasetname{}-Home, BLINK~\cite{fu2024blink}, and SpatialBench~\cite{cai2024spatialbot}.
\datasetname{}-Home contains 350 manually written spatial questions over diverse real-world RGBD scenes captured with an iPhone equipped with a depth sensor. We curated this benchmark to evaluate generalization to novel indoor settings with previously unseen objects.
BLINK is a visual reasoning benchmark consisting of binary spatial questions involving relationships such as ``next to,'' ``touching,'' and ``on top.'' 
BLINK allows us to assess the ability of models to assess generalization of spatial reasoning to unseen language configurations.
We evaluate only the \textit{spatial} portion of BLINK, as that aligns with the core focus of \datasetname{}.
SpatialBench, introduced in the SpatialBot~\cite{cai2024spatialbot} paper, uses RGB inputs and tests spatial understanding across several categories.
We focus on the \textit{position} category, which most directly aligns with our emphasis on spatial localization and placement.
Together, these benchmarks complement \datasetname{}-Val by covering a broader range of visual and linguistic variation.

\subsection{Results}

We evaluate the effectiveness of \datasetname{} in improving spatial reasoning capabilities in VLMs across held-out and out-of-domain benchmarks. In this section, we focus on analyzing the model’s generalization and understanding of spatial relationships. We address the following questions:

\noindent \textbf{How well does \datasetname{} training generalize to unseen spatial relationships?}
Although \datasetname{} consists of template-generated QA pairs with a fixed set of spatial prepositions, we observe in \cref{tab:downstream_results} that models trained on it can generalize to spatial relationships not explicitly included in the training set. 
This is particularly evident in evaluations on the BLINK dataset~\cite{fu2024blink}, which contains diverse prepositions such as “under,” “next to,” and “far away.”
We attribute this generalization to the fact that \datasetname{} encompasses all six principal directions in 3D space (along the x, y, and z axes). 
Generalizing to new prepositions often requires mapping linguistic expressions (\eg, “on top of,” “under”) to these spatial primitives—a task at which LLMs are naturally proficient. 
For example, “on top of” often refer to “above” in a world-centric frame, while “under” maps to “below.”
Moreover, prepositions such as “next to” or “beside” imply proximity between objects. 
Because \datasetname{} includes questions that require generating points near a reference object, it implicitly teaches the concept of closeness. 
This enables trained models to understand these proximity-based relationships, even if they are not explicitly represented during training. 

\noindent \textbf{Do \datasetname{}-trained models understand nuanced perspectives?}
Spatial references in natural language often imply specific reference frames. 
For instance, “in front of the car” typically refers to the direction of the car’s front hood. 
In \datasetname{}-Home, we omit explicit frame specifications in the questions to evaluate whether models can align with the implicit reference frame intended by the questioner.
We find that models trained with \datasetname{} can often infer the correct frame of reference, suggesting that they have learned to associate object geometries and orientations with spatial language. 
Figure~\ref{fig:exp} shows examples such as “Is the frame in front of the window?”, where the model accurately identifies the intended spatial relation.

\noindent \textbf{Are 3D VLMs better at learning spatial relationships than 2D VLMs?}
The findings in \cref{tab:main_results} suggest that 3D VLMs tend to outperform 2D counterparts in spatial reasoning tasks, likely due to their ability to directly utilize depth information. 
However, this comparison is not entirely fair: models like 3D-LLM~\cite{3dllm} and LEO~\cite{huang2023embodied} are pretrained on RGB-D indoor scan datasets, some of which overlap with the environments used in the source datasets (\eg, Matterport3D, ScanNet). 
This gives them prior exposure to scene geometry and object layouts, which may bias their performance.
To support more controlled and fair comparisons in the future, we designed \datasetname{} to be compatible with both 2D and 3D modalities, allowing researchers to investigate the impact of modality, architecture, and pretraining data under unified evaluation protocols.

\begin{table}
    \centering
    \begin{minipage}{0.52\linewidth}
    \centering
    \renewcommand\arraystretch{0.9}
    \resizebox{\linewidth}{!}{
    \begin{tabular}{lcc}
        \toprule
        \textbf{}              & \textbf{Indoor} & \textbf{Tabletop} \\ 
        \textbf{}              & \textbf{$\downarrow$} & \textbf{$\downarrow$} \\ 
        \textbf{}              & \textbf{Tabletop} & \textbf{Indoor} \\ 
        \midrule
        RoboPoint~\cite{yuan2024robopoint} & 38.7\,\,\,\,                                   & 38.2\,\,\,\,                                            \\ 
        \,\,\,+\datasetname{}  & 48.9~\grupa                                  & 51.3~\grupa                                        \\
         LEO~\cite{huang2023embodied}   &   41.9\,\,\,\,                                &    43.7\,\,\,\,                                          \\ 
         \,\,\,+\datasetname{} &   47.2~\grupa                          &    54.5~\grupa            \\
        \bottomrule
    \end{tabular}
    }
    \caption{Cross-dataset generalization results between indoor and tabletop environments.}
    \label{tab:dataset_generalization}
    \end{minipage}
    \hspace{0.02\linewidth}
    \begin{minipage}{0.42\linewidth}
    \centering
    \renewcommand\arraystretch{0.9}
    \resizebox{\linewidth}{!}{
    \begin{tabular}{l@{\hspace{0.5em}}r}
        \toprule
        \textbf{Model} & \textbf{Success (\%)} \\
        \midrule
        \multicolumn{2}{c}{\textit{Open-source}} \\
        LLaVA-NeXT~\cite{liu2023improvedllava} & 23.7\,\,\,\, \\
        \,\,+~\datasetname{} & \textbf{52.6}~\grupa \\
        RoboPoint~\cite{yuan2024robopoint} & 44.7\,\,\,\, \\
        \,\,+~\datasetname{} & 46.2~\grupa \\
        \midrule
        \multicolumn{2}{c}{\textit{Not available for fine-tuning}} \\
        Molmo~\cite{deitke2024molmopixmoopenweights} & 43.8\,\,\,\, \\
        GPT-4o~\cite{openai2022chatgpt} & 46.9\,\,\,\, \\
        \bottomrule
    \end{tabular}
    }
    \caption{Robot experiment results.}
    \label{tab:robot_results}
    \end{minipage}
    \end{table}

\subsection{Real Robot Experiments}

\begin{figure}
    \vspace{-3.5mm}
    \centering
    \includegraphics[width=\linewidth]{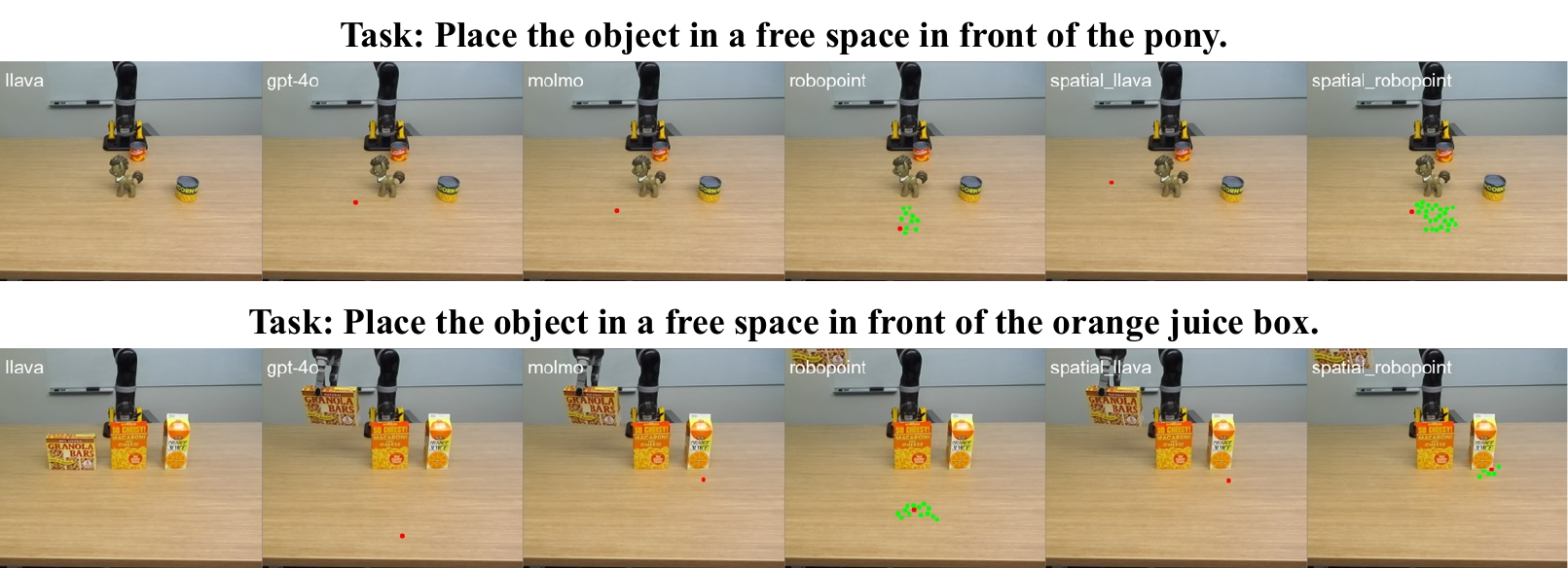}
    \vspace{-3mm}
    \caption{Robotics experiments: the red dot shows the model output (if not present, the model failed to provide a valid point in the image);  green dots are used to show when a model outputs multiple points. The robot motion generator, cuRobo~\cite{sundaralingam2023curobo}, is used to grasp the item referenced by the generated point. The \textit{spatial-} prefix indicates model trained with \datasetname{}.}
    \label{fig:robot_experiment}
        \vskip -10pt
\end{figure}

We design a suite of tabletop manipulation tasks requiring spatial reasoning. The setup includes a Kinova Jaco robot~\cite{kinova}, paired with a ZED2 camera for RGB-D perception.
The robot system implements actions to \emph{pick} and \emph{place} objects on the table using  cuRobo~\cite{sundaralingam2023curobo} for motion planning. 
Tasks include spatial questions that require a yes/no answer, and pick-and-place instructions that require successfully controlling the robot to complete the task.
We adopt a modular design, where the VLM is queried for spatial understanding, and the resulting predictions (\eg, target points) are passed to a separate motion planning system for execution.
We use a range of simple, unambiguous objects—colored cubes, cylinders, food items, and toys—to ensure the challenge lies in spatial understanding rather than object recognition (Figure~\ref{fig:robot_experiment}). 
In total, we conducted over 200 model queries. Details of the questions and scene configurations are provided in the Appendix~\ref{sec:appendix:robot}.
We evaluate the following VLMs: LLaVA-NeXT~\cite{liu2023improvedllava} and RoboPoint~\cite{yuan2024robopoint}, both with and without \datasetname{} training; and two strong baselines, Molmo~\cite{deitke2024molmopixmoopenweights} and GPT-4o~\cite{openai2022chatgpt}. Table~\ref{tab:robot_results} and Figure~\ref{fig:robot_experiment} present the results.

Experiments show that LLaVA-NeXT fine-tuned on \datasetname{} achieves the highest success rate across all models. 
Training with \datasetname{} enhances spatial understanding in 2D VLMs, enabling the model to correctly interpret instructions such as “place in front of the pony,” where placement is aligned with the pony’s head direction. 
It also demonstrates sensitivity to object scale, as in the task “place in front of the orange juice box,” where the model places the object at a reasonable distance. 
In contrast, baseline models such as RoboPoint frequently place objects too far from the target, likely due to limited understanding of spatial proximity. 
We also observe that spatial failures in 2D VLMs often stem from errors in projecting 2D predictions into 3D. 
Even a small 2-pixel shift in image space can translate to a 5–10 cm error in the physical world, which is significant in manipulation tasks. 
Nonetheless, models trained on \datasetname{} produce more accurate predictions, reducing these failure cases and showing the benefit of dataset-driven improvements.
Interestingly, GPT-4o performs comparably to \datasetname{}-trained RoboPoint. 
We attribute this to GPT-4o’s broader language understanding and instruction-following ability, which partially compensates for its lack of task-specific spatial training. 
Looking forward, promising directions include investigating how viewpoint affects 2D spatial predictions, and developing 3D VLMs that can reason over partial point clouds—removing the need for complete 3D scans and making deployment in real-world systems more feasible.

\section{Conclusion}
We introduce \datasetname{}, \datasetname{}-Val, and \datasetname{}-Home, a large-scale 2D/3D spatial understanding training and evaluation dataset tailored for robotics. 
Experimental results show that models trained with \datasetname{} are able to understand spatial relationships, generalize to unseen relationships, and infer nuanced reference frames, making them applicable in a wide range of tasks that require spatial understanding.
We further demonstrate the real-world applicability of \datasetname{} with robot experiments. 
In addition, our automatic data generation pipeline can be used to extend the dataset to new data sources and spatial relations.
We show that \datasetname{} has the potential to serve as a foundation for broader applications in robotics which require spatial understanding.

\section*{Acknowledgements}

The authors thank Youngsun Wi, Hyunho Ahn, Hojin Yoo, and Minjae Bae for providing images in the \datasetname{}-Home dataset.
Our research was supported in part by ARL W911NF2220144 and resources from the Ohio Supercomputer Center. 
These research findings were partially derived using the Matterport dataset, the use of which is governed by: \small{\url{https://kaldir.vc.in.tum.de/matterport/MP\_TOS.pdf}}.

{
    \small
    \bibliographystyle{ieeenat_fullname}
    \bibliography{main}
}

\clearpage

\appendix

\section*{Appendices}
\label{sec:Appendix}

In this supplementary material, we present additional details and clarifications that are omitted in the main text due to space constraints.
\begin{itemize}
    \item \autoref{supp:limitations} Limitations.
    \item \autoref{supp:dataset} Dataset Details.
    \item \autoref{supp:implementation} Implementation Details.
    \item \autoref{supp:result} More Results.
\end{itemize}

\section{Limitations}
\label{supp:limitations}
While \datasetname{} significantly improves spatial reasoning capabilities in VLMs, certain design choices naturally introduce trade-offs and areas for future exploration.

First, the dataset relies on a top-down occupancy map to identify and annotate empty regions for spatial context and compatibility tasks. 
This approach simplifies reasoning about object placement on horizontal surfaces and enables efficient data generation, but it currently does not support spatial questions involving containment—such as whether an object can fit inside or under another object—which would require more detailed volumetric modeling.

Second, although the models are deployed on a real robot using a modular approach, we do not yet explore tighter forms of integration such as training it jointly with robot trajectories~\cite{kim24openvla}.
Investigating these alternatives could enhance downstream policy learning and enable more seamless end-to-end systems.

Finally, \datasetname{} focuses on indoor and tabletop scenes containing objects commonly encountered in household environments, and does not include humans or animals. 
This reflects the nature of source datasets and our emphasis on robot object manipulation. 
While this limits coverage of social or dynamic interaction scenarios, trained models still generalizes well to out-of-distribution benchmarks like BLINK, which include humans and animals—suggesting that the learned spatial representations are broadly transferable.

\section{Dataset Details}
\label{supp:dataset}

\subsection{Dataset Statistics} We provide the full dataset statistics in~\cref{tab:supp_datasets}. 
For all training, we use only 900,000 spatial relationships, sampled equally across all datasets, due to computational constraints.
We further experiment on the effect of data scaling on \cref{tab:supp_scaling_annotation} and explain the results.
Notably, HOPE~\cite{tyree2022hope} and GraspNet-1B~\cite{fang2020graspnet} contain similar tabletop images captured from different perspectives, resulting in lower dataset diversity for the tabletop environment.
We plan to enhance the diversity of \datasetname{} by incorporating additional tabletop datasets.

\subsection{Choice of Spatial Relationships} 
In designing the dataset, we focused on spatial relationships that directly impact robotic perception, planning, and interaction: context, compatibility, and configuration. 
These were selected to reflect the core spatial reasoning challenges that robots encounter when operating in complex, real-world environments. 

We intentionally excluded tasks such as object counting, as we consider them to fall outside the scope of spatial understanding. 
While counting is an important visual reasoning skill, it does not require reasoning about spatial relations between objects or between objects and their environment. 
For example, determining that “three cups are on the table” is a perceptual task rather than a spatial reasoning one. 
As such, counting may complement but does not substitute for the types of relational reasoning we target. 
We leave the integration of counting tasks into spatial benchmarks as future work.

Similarly, we exclude tasks that rely solely on distance measurements. 
Although distance is a fundamental spatial quantity, it is difficult to define consistently across different environments, object scales, and robot embodiments. 
Absolute distances can vary significantly between indoor and outdoor scenes, small and large objects, or different robot perspectives, making them hard to normalize or interpret in a general way. 
Moreover, distance alone often lacks the relational semantics required for higher-level reasoning—for example, understanding that an object is behind, above, or in front of others. 
\datasetname{} instead focuses on spatial relationships that are more invariant, interpretable, and transferable across diverse robotic scenarios.

That said, the data generation pipeline is general and could readily support auxiliary tasks involving object counting or distance estimation if desired. 
These metrics may serve as useful complements in future extensions of the benchmark or as auxiliary supervision signals in model training.

\subsection{Object Grounding Dataset}
\label{appendix:object_grounding}

To support accurate spatial understanding, we generate an auxiliary dataset for object grounding. 
Many spatial reasoning tasks assume that the model can correctly identify which object is being referred to in the scene. 
However, in practice, this can be a major source of error—especially in cluttered environments or when multiple instances of the similar object type are present.

The grounding dataset provides direct supervision to help models learn to associate text descriptions with specific objects in the image. 
For each image, we include a set of object descriptions (\eg, ``the keyboard'' or ``the chair'') paired with the corresponding 2D bounding box of the object in the image. 
These 2D boxes are projected from the annotated 3D bounding boxes using camera intrinsics and extrinsics.

A total of 100k grounding QA pairs are generated and used during training to reduce reference ambiguity and improve object identification accuracy in spatial tasks.
While not part of the main spatial reasoning taxonomy, grounding accuracy is a prerequisite for answering spatial questions correctly, and we find that including this data helps reduce errors caused by incorrect object identification.

\begin{figure}[tb] \centering
    \includegraphics[width=\linewidth]{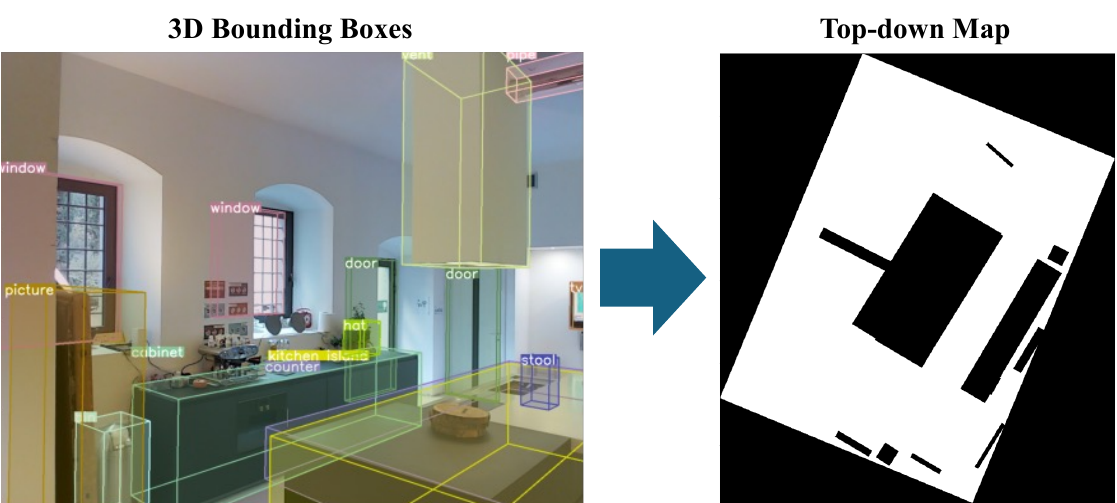}
    \caption{An example of generated top-down map of the image from 3D bounding boxes.} \label{fig:supp_topdown}
\end{figure}

\begin{table*}[ht]
\centering
\resizebox{\linewidth}{!}{
\begin{tabular}{lcccccccc}
\toprule
\textbf{Category}  & \textbf{Dataset}         & \textbf{Split}   & \textbf{Scans} & \textbf{Images} & \textbf{Configuration Q} & \textbf{Context Q} & \textbf{Compatibility Q} \\ 
\midrule
\multirow{6}{*}{Indoor} 
    & \multirow{2}{*}{Matterport3D~\cite{chang2017matterport3d}} & Train        & 1859 scans      & 236243   &  298439      &  298439  &  298439    \\ 
    &                               & Validation         & 10 scans        & 200       &  200   & 200    & 200     \\ 
    \cmidrule{2-8}
    & \multirow{2}{*}{ScanNet~\cite{dai2017scannet}}      & Train        & 1514 scans      & 280402     &  299039      &  299039  &  299039   \\ 
    &                               & Validation         &  12 scans        & 400       & 400   & 400    & 400     \\ 
    \cmidrule{2-8}
    & \multirow{2}{*}{3RScan~\cite{Wald2019RIO}}       & Train        & 1543 scans      & 366755   &  298839      &  298839  &  298839  \\ 
    &                               & Validation         & 18 scans        & 400       & 400   & 400    & 400    \\ 
\midrule
\multirow{4}{*}{Tabletop} 
    & \multirow{2}{*}{HOPE~\cite{tyree2022hope}}         & Train        & 60 scenes      & 50050     & 36817 & 36817    &  36817    \\  
    &                               & Validation         & 47 scenes       & 235       & 500   & 500    & 500    \\ 
    \cmidrule{2-8}
    & \multirow{2}{*}{GraspNet-1B~\cite{fang2020graspnet}}  & Train        & 130 scenes      & 25620      & 36817 & 36817    &  36817    \\ 
    &                               & Validation         & 30 scenes       & 120       & 500   & 500    & 500   \\  
\bottomrule
\end{tabular}
}
\caption{Full dataset statistics for indoor and tabletop datasets.}
\label{tab:supp_datasets}
\end{table*}

\subsection{Dataset Generation Details}

The dataset generation pipeline is detailed in the main text (\autoref{sec:data_gen}), which introduces a two-stage process for computing 3D spatial relationships and projecting them into 2D image space.
Here, we expand on implementation details not covered in the main paper and provide clarification on the reasoning logic used in spatial annotation.

\noindent \textbf{Reference Frame Annotation.}
For each spatial configuration question, we label relationships from three perspectives: ego-centric (camera view), object-centric (based on object heading), and world-centric (aligned with the dataset’s global frame).
To compute object-centric directions, we use the heading vector of each oriented 3D bounding box to define the “front” of the object. Left, right, behind, and front relations are then assigned accordingly.
World-centric annotations modify vertical relationships (above/below) using global $z$-coordinates to reflect elevation.

\noindent \textbf{Surface Detection and Free Space Sampling.}
To identify support surfaces such as tables, counters, or floors, we use GPT-4o to select candidate objects that are likely to support placement.
A top-down occupancy map is constructed from bounding boxes in the scene ~\cref{fig:supp_topdown}. We sample 3D points in unoccupied regions and project them into the image plane for spatial context tasks.
Points are filtered via occlusion checks using raycasting, ensuring sampled points are visible and unobstructed.

\noindent \textbf{Compatibility Check and Object Placement.}
For spatial compatibility, we simulate placing a virtual object bounding box at candidate locations.
The placement must fit without intersecting other objects and must allow a clearance of at least 10~cm in all axes.
We allow in-plane rotation and translation to test flexible placement.
This provides a binary label (True/False) indicating whether the object can be compatibly placed in the region.

\noindent \textbf{Output Format.}
Though \datasetname{} uses point prediction for ease of integration with robot setups, the pipeline also supports mask-based outputs and can be extended in future work.


\section{Implementation Details}
\label{supp:implementation}

\begin{table*}[ht]
    \centering
    
    \tabcolsep 5pt
    \renewcommand\arraystretch{1}
    \captionsetup{width=0.9\textwidth} 
    \resizebox{1\textwidth}{!}{
    \begin{tabular}{lccccccccc}
    \toprule
    \multirow{2}{*}{\textbf{Model}} &  \multicolumn{3}{c}{\textbf{Indoor}} & \multicolumn{3}{c}{\textbf{Tabletop}} & \multicolumn{3}{c}{\textbf{Average}}\\
     \cmidrule(r){2-4} \cmidrule(r){5-7} \cmidrule(r){8-10} &  \textbf{Ego-centric}  & \textbf{Object-centric} & \textbf{World-centric} &  \textbf{Ego-centric}  & \textbf{Object-centric} & \textbf{World-centric} & \textbf{Indoor}  & \textbf{Tabletop}  &\textbf{Total} \\
    \midrule 
    \multicolumn{10}{c}{\textit{Open-source VLMs}} \\
    \multicolumn{10}{c}{\textbf{2D VLMs}} \\
    VILA~\cite{lin2023vila} & 55.9\,\,\,\, & 40.5\,\,\,\,  & 32.9\,\,\,\, &  43.6\,\,\,\, & 39.7\,\,\,\, & 28.9\,\,\,\, &  43.1\,\,\,\, & 37.4\,\,\,\, &  40.2\,\,\,\,   \\
    \,\,\,+\datasetname{} &  74.3\grupa & 57.8~\grupa &  62.3~\grupa & 70.3~\grupa & 58.1~\grupa &  60.3~\grupa  & 64.8~\grupa & 62.9~\grupa & 63.9~\grupa \\
    \cmidrule(r){1-1}
    LLaVA-Next~\cite{liu2023improvedllava} & 35.2\,\,\,\,  & 24.3\,\,\,\, &  34.7\,\,\,\, & 36.4\,\,\,\, & 28.5\,\,\,\, & 22.7\,\,\,\,  & 31.4\,\,\,\,  & 29.2\,\,\,\, &   30.3\,\,\,\,  \\
    \,\,\,+\datasetname{} &  75.4~\grupa & 54.1~\grupa &  68.8~\grupa & 67.9~\grupa & 54.7~\grupa &  58.9~\grupa  & 60.4~\grupa & 60.5~\grupa & 60.5~\grupa  \\
    \cmidrule(r){1-1}
    SpaceLLaVA~\cite{chen2024spatialvlm} & 40.6\,\,\,\,  & 36.0\,\,\,\, &  30.1\,\,\,\, & 52.3\,\,\,\, &  32.8 \,\,\,\, & 53.5\,\,\,\,  &  38.9\,\,\,\, & 46.2\,\,\,\, & 43.6\,\,\,\,  \\
    \,\,\,+\datasetname{} &  \textbf{78.5}~\grupa & \textbf{60.6}~\grupa &  64.3~\grupa & 73.0~\grupa & 49.5~\grupa &   68.3~\grupa  &  67.8~\grupa & 63.6~\grupa & 65.7~\grupa \\
    \cmidrule(r){1-1}
    RoboPoint~\cite{yuan2024robopoint} &  41.9\,\,\,\, & 36.2\,\,\,\, & 40.7\,\,\,\,  &  46.2\,\,\,\, &  30.5\,\,\,\, & 37.9\,\,\,\, &  39.6\,\,\,\, & 38.2\,\,\,\, & 38.9\,\,\,\, \\
    \,\,\,+\datasetname{} &  76.4~\grupa & 58.3~\grupa &  78.3~\grupa &  \textbf{76.7}~\grupa & 62.6~\grupa &  71.0~\grupa & 71.0~\grupa & 70.1~\grupa & 70.6~\grupa  \\ \\

    \multicolumn{10}{c}{\textbf{3D VLMs}} \\
    3D-LLM~\cite{3dllm} & 28.9\,\,\,\, & 38.3\,\,\,\, & 45.6\,\,\,\,  &  38.9\,\,\,\, &  35.7\,\,\,\, &  52.6\,\,\,\, &  37.6\,\,\,\, & 42.4\,\,\,\, &  40.0\,\,\,\, \\ 
        \,\,\,+\datasetname{} &  60.7~\grupa &  52.1~\grupa &  76.5~\grupa & 57.9~\grupa &  \textbf{62.8}~\grupa &  77.3~\grupa &  63.1~\grupa & 66.0~\grupa & 64.6~\grupa \\ 
    \cmidrule(r){1-1}
    LEO~\cite{huang2023embodied} &  46.9\,\,\,\,  & 30.6\,\,\,\, &  48.2\,\,\,\,  &  41.4\,\,\,\, &  34.3\,\,\,\, &   55.4\,\,\,\, &  41.9\,\,\,\, &  43.7\,\,\,\,  & 42.8\,\,\,\, \\
    \,\,\,+\datasetname{} &  68.1~\grupa &  71.6~\grupa &  \textbf{79.6}~\grupa & 71.4~\grupa &  60.2~\grupa &  \textbf{80.5}~\grupa & \textbf{73.1}~\grupa  & \textbf{70.7}~\grupa &  \textbf{71.9}~\grupa \\ 
    
    \midrule
    \multicolumn{10}{c}{\textit{Not available for fine-tuning}} \\
    \multicolumn{10}{c}{\textbf{2D VLMs}} \\
    Molmo~\cite{deitke2024molmopixmoopenweights} & 50.4\,\,\,\, & 50.8\,\,\,\, & 47.6\,\,\,\, & 64.4\,\,\,\, & 33.6\,\,\,\, & 53.8\,\,\,\, & 49.6\,\,\,\,  & 50.6\,\,\,\, & 50.1\,\,\,\, \\
    GPT-4o~\cite{openai2022chatgpt} & 52.9\,\,\,\,  & 38.7\,\,\,\, & 56.3\,\,\,\,  &  62.5\,\,\,\, & 30.7\,\,\,\, & 63.7\,\,\,\, & 49.3\,\,\,\,  & 52.3\,\,\,\, & 50.8\,\,\,\, \\ 
    \bottomrule
    \end{tabular}
    }
    \caption{Results of per frame accuracy of existing 2D/3D VLMs on a \datasetname{}-Val.  All methods, for all tasks, perform better (\grupa) when fine-tuned on \datasetname{}. The best result for each column is bolded.}

    \label{tab:supp_frame_main_results}
\end{table*}

\begin{table}[t]
    \centering
    \resizebox{\linewidth}{!}{
    \begin{tabular}{lccccc}
        \toprule
          & \textbf{100K} & \textbf{300K} & \textbf{900k} (Default) & \textbf{1.8M} & \textbf{3M} (Full)  \\
         \midrule
         LLaVa-Next~\cite{liu2023improvedllava} & 38.1  &  46.7  & 60.5 & 65.8 & 72.4   \\
         \bottomrule
    \end{tabular}
    }
    \caption{Results of scaling experiment on LLaVa-Next~\cite{liu2023improvedllava} with varied number of spatial relationship annotations. Average accuracy on \datasetname-Val is reported.}
    \label{tab:supp_scaling_annotation}
\end{table}

\begin{table}[t]

    \centering
    \small
    \renewcommand\arraystretch{0.8}
    \resizebox{\linewidth}{!}{
    \begin{tabular}{lcccc}
        \toprule
        \textbf{}               & \textbf{MMMU$_{val}$} & \textbf{MME$_p$} & \textbf{MME$_c$} & \textbf{MMBench$_{dev}$}\\ 
        \midrule
        LLaVA-NeXT & 39.4  &  1561.8 & \textbf{305.4}  &   \textbf{71.6}                                     \\ 
        +\datasetname{}  &  \textbf{39.8}    &  \textbf{1604.5}  &   293.2   &  \textbf{71.6}                                                              \\
        
         
        \bottomrule
    \end{tabular}
    }
    \caption{Evaluation on general-purpose multimodal benchmarks (MMMU, MME, MMBench) to assess whether training on \datasetname{} affects commonsense and factual reasoning.}
    \label{tab:rebuttal_dataset_generalization_mm}
\end{table}

\begin{table}[t]

    \centering
    \small
    \renewcommand\arraystretch{0.5}
    \resizebox{\linewidth}{!}{
    \begin{tabular}{lcccc}
        \toprule
        \textbf{}              & \textbf{Base} & \textbf{Auxiliary} & \textbf{\datasetname{}} & \textbf{Both} \\ 
        \midrule
        LLaVA-NeXT 
        & 30.3 &        32.4           &   51.8             &        60.5 \\                               
        \bottomrule
    \end{tabular}
    }
    \caption{Ablation study evaluating the impact of the auxiliary grounding dataset on \datasetname{}-Val. }
    \label{tab:rebuttal_grounding}
\end{table}

\subsection{Model Training}
We further explain the training details for all 2D and 3D VLMs trained on \datasetname{}. 
For all models, we perform instruction tuning using the model weights from public repositories. 
All training is done using 8 Nvidia H100 GPUs, with the training time between 20 and 40 hours.

\subsection{Model Setup}

\noindent \textbf{VILA}~\cite{lin2023vila} We initialize the model from Efficient-Large-Model/Llama-3-VILA1.5-8B on Hugging Face. 
We use the fine-tuning script from the VILA GitHub repository to train the model using the default hyperparameters. \\
\noindent \textbf{LLaVA-NeXT}~\cite{liu2023improvedllava} We initialize the model from lmms-lab/llama3-llava-next-8b on Hugging Face. We use the LLaVA-Next fine-tuning script from the LLaVA-Next repository using the default hyperparameters. \\
\noindent \textbf{SpaceLLaVA}~\cite{chen2024spatialvlm} 
As official code and weights for SpatialVLM~\cite{chen2024spatialvlm} is not released, we use a community implementation which is endorsed by SpatialVLM~\cite{chen2024spatialvlm} authors.
We initialize the model from remyxai/SpaceLLaVA from Hugging Face. We use LLaVA-1.5 finetuning script from LLaVa~\cite{liu2023llava} repository using the default hyperparameters. \\
\noindent \textbf{RoboPoint}~\cite{yuan2024robopoint} We initialize the model from wentao-yuan/robopoint-v1-vicuna-v1.5-13b on Hugging Face. 
We use the fine-tuning script provided in the RoboPoint~\cite{yuan2024robopoint} GitHub repository to train the model using the default hyperparameters. \\
\noindent \textbf{3D-LLM}~\cite{3dllm} We initialize the model using the pretrain\_blip2\_sam\_flant5xl\_v2.pth checkpoint downloaded from the official GitHub repository. 
Since the model requires preprocessing of multiview images, we follow the author's pipeline to process multiview images from the environments.
Because the model does not accept image input, we append the following text in front of the question to ensure the model understands the perspective from which the question is being asked: ``I am facing {\textsc{anchor object}}." We use the default hyperparameters and train the model for 20 epochs per the author's guidelines. 
We choose the best model based on validation accuracy. \\
\noindent \textbf{LEO}~\cite{huang2023embodied} We initialize the model from the sft\_noact.pth checkpoint downloaded from the official GitHub repository. \\
Since LEO supports dual image and 3D point cloud input, we input both of them and modify the question as in 3D-LLM. 
We use the default hyperparameters and train the model for 10 epochs per the author's guidelines, and choose the best model based on validation accuracy.

We could not fine-tune Molmo~\cite{deitke2024molmopixmoopenweights} from allenai/Molmo-7B-D-0924 or GPT-4o~\cite{openai2022chatgpt} from the gpt-4o-2024-08-06 API due to the unavailability of the fine-tuning script at the time of this work, thus we use them as a zero-shot baselines.

\section{More Results}
\label{supp:result}

\subsection{Accuracy Per Reference Frame}

We show the results per frame in~\cref{tab:supp_frame_main_results} for \datasetname-Val.
From the results, we can see a distinct difference between 2D and 3D VLMs in understanding the world-centric frame before training with \datasetname{}.
Baseline 2D VLMs have trouble understanding the world-centric frame, which involves understanding elevation, while 3D VLMs comparatively excel at it.
Furthermore, we can see that since baseline 3D VLMs are trained on point clouds without information of perspective, their accuracy in ego-centric and object-centric frames is lower.
However, with \datasetname{} training, we were able to teach the 3D VLMs to think in a certain frame, thus considerably improving their performance on ego-centric and object-centric frames.
However, we hypothesize that, due to their design—specifically, the lack of a means to visually inject perspective information since they require complete 3D point clouds—3D VLMs still lag behind 2D VLMs on ego-centric and object-centric frames.

\subsection{Data Scaling}

In \cref{tab:supp_scaling_annotation}, we experiment with scaling the number of annotations while keeping images fixed. 
We found that even though the number of images stays consistent, increasing the number of annotations can improve performance. 
For future work, we plan to apply the data generation pipeline to a diverse set of indoor and tabletop environments to further improve the performance of the models.

\subsection{Commonsense Knowledge Retention}
To ensure that training on \datasetname{} does not degrade a model’s general reasoning or commonsense capabilities, we evaluate the RoboSpatial-trained model on a suite of standard multimodal benchmarks: MMMU~\cite{yue2023mmmu}, MME~\cite{fu2023mme}, and MMBench~\cite{MMBench}. 
As shown in Table~\ref{tab:rebuttal_dataset_generalization_mm}, the \datasetname{}-trained model maintains or slightly improves performance across all benchmarks, suggesting that spatial fine-tuning preserves broader knowledge capabilities.

\subsection{Ablation of the Auxiliary Grounding Dataset}


As shown in Table~\ref{tab:rebuttal_grounding}, training on the auxiliary dataset alone yields a small improvement over the base model (+2.1), but it falls far short of the gains achieved with \datasetname{}, which is explicitly designed to teach spatial reasoning.
This confirms that grounding supervision alone is insufficient for spatial understanding.
However, combining both datasets leads to the best performance, suggesting that improving object localization can complement spatial supervision when jointly trained.

\subsection{Robot Experiments Details}
\label{sec:appendix:robot}

\subsubsection{Robot Setup}

For picking, we find which object the point maps to using SAM 2~\cite{ravi2024sam2} and execute the picking behavior on that object. 
For placing, we simply compute the 3D coordinate based on the depth value at that pixel and place the object at that coordinate. 
There were no failures due to cuRobo~\cite{sundaralingam2023curobo} failing. 
The experiments were purposely designed to consist of behaviors that our robot system can handle in order to avoid introducing irrelevant factors. 
The picking behavior consists of computing a top-down grasp pose and reaching it with cuRobo~\cite{sundaralingam2023curobo}. To compute the grasp pose: 
\begin{enumerate} 
    \item We estimate the major axis of the object's point cloud in top-down view using PCA. 
    \item The grasp orientation is orthogonal to the major axis. 
    \item The grasp height is based on the highest point in the object's point cloud minus an offset of 3cm. This heuristic ensures the system can grip long objects. 
\end{enumerate} 
The placing behavior is the same as picking, except that an area within 5cm of the placement coordinate is used as the point cloud for estimating orientation and height, and a vertical height offset is added to account for the height at which the object was picked.

\subsubsection{Additional Results}

\begin{figure*}[ht]
\centering
\captionsetup{width=0.9\textwidth} 
\begin{scriptsize}

\begin{minipage}{\linewidth}
\centering

\begin{minipage}{0.48\linewidth}
\centering
\begin{minipage}{0.4\linewidth}
\centering
\includegraphics[width=\textwidth]{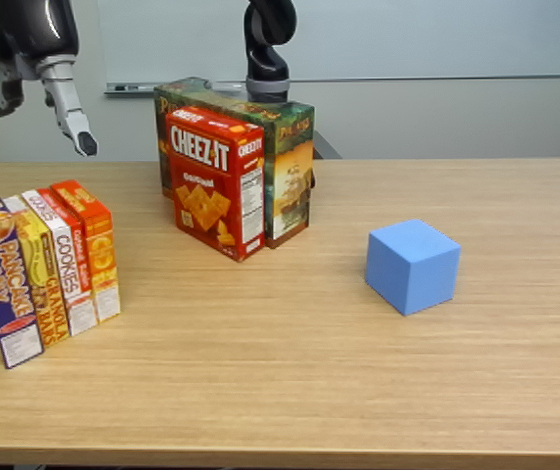}
\end{minipage}%
\hfill
\begin{minipage}{0.58\linewidth}
\centering
\begin{tabular}{p{0.85\linewidth}c}
\toprule
\multicolumn{2}{p{\linewidth}}{\textbf{Question:} pick lone object} \\
\midrule
LLaVa-Next~\cite{liu2023improvedllava} & \textcolor{red}{\texttimes} \\
LLaVa-Next\textsc{-FT}~\cite{liu2023improvedllava} & \textcolor{green}{\checkmark} \\
RoboPoint~\cite{yuan2024robopoint} & \textcolor{red}{\texttimes} \\
RoboPoint\textsc{-FT}~\cite{yuan2024robopoint} & \textcolor{green}{\checkmark} \\
Molmo~\cite{deitke2024molmopixmoopenweights} & \textcolor{green}{\checkmark} \\
GPT-4o~\cite{openai2022chatgpt} & \textcolor{red}{\texttimes} \\
\bottomrule
\end{tabular}
\end{minipage}
\end{minipage}%
\hfill
\begin{minipage}{0.48\linewidth}
\centering
\begin{minipage}{0.4\linewidth}
\centering
\includegraphics[width=\textwidth]{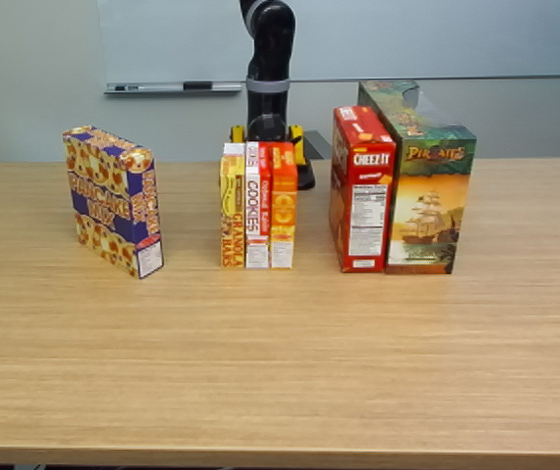}
\end{minipage}%
\hfill
\begin{minipage}{0.58\linewidth}
\centering
\begin{tabular}{p{0.85\linewidth}c}
\toprule
\multicolumn{2}{p{\linewidth}}{\textbf{Question:} Is there room to slot the pancake mix in the middle of the row of boxes} \\
\midrule
LLaVa-Next~\cite{liu2023improvedllava} & \textcolor{green}{\checkmark} \\
LLaVa-Next\textsc{-FT}~\cite{liu2023improvedllava} & \textcolor{green}{\checkmark} \\
RoboPoint~\cite{yuan2024robopoint} & \textcolor{red}{\texttimes} \\
RoboPoint\textsc{-FT}~\cite{yuan2024robopoint} & \textcolor{green}{\checkmark} \\
Molmo~\cite{deitke2024molmopixmoopenweights} & \textcolor{green}{\checkmark} \\
GPT-4o~\cite{openai2022chatgpt} & \textcolor{green}{\checkmark} \\
\bottomrule
\end{tabular}
\end{minipage}
\end{minipage}

\end{minipage}

\vspace{0.3cm}

\begin{minipage}{\linewidth}
\centering

\begin{minipage}{0.48\linewidth}
\centering
\begin{minipage}{0.4\linewidth}
\centering
\includegraphics[width=\textwidth]{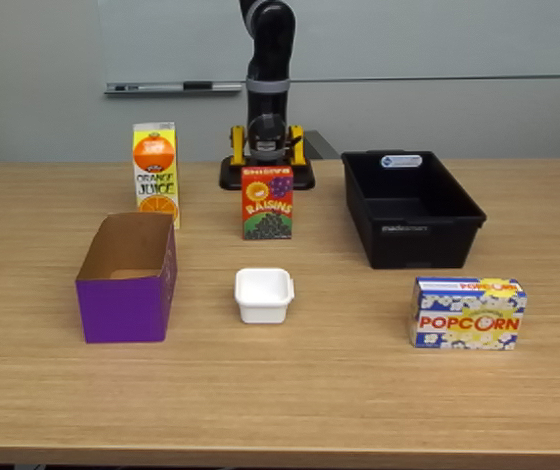}

\end{minipage}%
\hfill
\begin{minipage}{0.58\linewidth}
\centering
\begin{tabular}{p{0.85\linewidth}c}
\toprule
\multicolumn{2}{p{\linewidth}}{\textbf{Question:} Is there space in the white container for the orange juice box} \\
\midrule
LLaVa-Next~\cite{liu2023improvedllava} & \textcolor{red}{\texttimes} \\
LLaVa-Next\textsc{-FT}~\cite{liu2023improvedllava} & \textcolor{green}{\checkmark} \\
RoboPoint~\cite{yuan2024robopoint} & \textcolor{red}{\texttimes} \\
RoboPoint\textsc{-FT}~\cite{yuan2024robopoint} & \textcolor{red}{\texttimes} \\
Molmo~\cite{deitke2024molmopixmoopenweights} & \textcolor{red}{\texttimes} \\
GPT-4o~\cite{openai2022chatgpt} & \textcolor{green}{\checkmark} \\
\bottomrule
\end{tabular}
\end{minipage}

\vspace{0.3cm}

\begin{minipage}{0.4\linewidth}
\centering
\includegraphics[width=\textwidth]{supp_robot/experiments/8.png}

\end{minipage}%
\hfill
\begin{minipage}{0.58\linewidth}
\begin{tabular}{p{0.85\linewidth}c}
\toprule
\multicolumn{2}{p{\linewidth}}{\textbf{Question:} pick object behind the middle container} \\
\midrule
LLaVa-Next~\cite{liu2023improvedllava} & \textcolor{red}{\texttimes} \\
LLaVa-Next\textsc{-FT}~\cite{liu2023improvedllava} & \textcolor{green}{\checkmark} \\
RoboPoint~\cite{yuan2024robopoint} & \textcolor{green}{\checkmark} \\
RoboPoint\textsc{-FT}~\cite{yuan2024robopoint} & \textcolor{red}{\texttimes} \\
Molmo~\cite{deitke2024molmopixmoopenweights} & \textcolor{red}{\texttimes} \\
GPT-4o~\cite{openai2022chatgpt} & \textcolor{red}{\texttimes} \\
\bottomrule
\end{tabular}
\end{minipage}

\vspace{0.3cm}

\begin{minipage}{0.4\linewidth}
\centering
\includegraphics[width=\textwidth]{supp_robot/experiments/8.png}
\end{minipage}%
\hfill
\begin{minipage}{0.58\linewidth}
\begin{tabular}{p{0.85\linewidth}c}
\toprule
\multicolumn{2}{p{\linewidth}}{\textbf{Question:} place object in container behind popcorn} \\
\midrule
LLaVa-Next~\cite{liu2023improvedllava} & \textcolor{red}{\texttimes} \\
LLaVa-Next\textsc{-FT}~\cite{liu2023improvedllava} & \textcolor{green}{\checkmark} \\
RoboPoint~\cite{yuan2024robopoint} & \textcolor{green}{\checkmark} \\
RoboPoint\textsc{-FT}~\cite{yuan2024robopoint} & \textcolor{green}{\checkmark} \\
Molmo~\cite{deitke2024molmopixmoopenweights} & \textcolor{red}{\texttimes} \\
GPT-4o~\cite{openai2022chatgpt} & \textcolor{red}{\texttimes} \\
\bottomrule
\end{tabular}
\end{minipage}
\end{minipage}%
\hfill
\begin{minipage}{0.48\linewidth}
\centering
\begin{minipage}{0.4\linewidth}
\centering
\includegraphics[width=\textwidth]{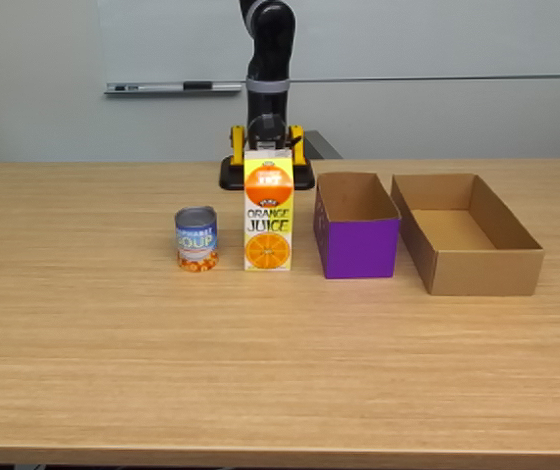}
\end{minipage}%
\hfill
\begin{minipage}{0.58\linewidth}
\centering
\begin{tabular}{p{0.85\linewidth}c}
\toprule
\multicolumn{2}{p{\linewidth}}{\textbf{Question:} alphabet soup fit in the purple box} \\
\midrule
LLaVa-Next~\cite{liu2023improvedllava} & \textcolor{green}{\checkmark} \\
LLaVa-Next\textsc{-FT}~\cite{liu2023improvedllava} & \textcolor{red}{\texttimes} \\
RoboPoint~\cite{yuan2024robopoint} & \textcolor{green}{\checkmark} \\
RoboPoint\textsc{-FT}~\cite{yuan2024robopoint} & \textcolor{green}{\checkmark} \\
Molmo~\cite{deitke2024molmopixmoopenweights} & \textcolor{red}{\texttimes} \\
GPT-4o~\cite{openai2022chatgpt} & \textcolor{green}{\checkmark} \\
\bottomrule
\end{tabular}
\end{minipage}

\vspace{0.3cm}

\begin{minipage}{0.4\linewidth}
\centering
\includegraphics[width=\textwidth]{supp_robot/experiments/9.png}
\end{minipage}%
\hfill
\begin{minipage}{0.58\linewidth}
\begin{tabular}{p{0.85\linewidth}c}
\toprule
\multicolumn{2}{p{\linewidth}}{\textbf{Question:} pick shortest object} \\
\midrule
LLaVa-Next~\cite{liu2023improvedllava} & \textcolor{red}{\texttimes} \\
LLaVa-Next\textsc{-FT}~\cite{liu2023improvedllava} & \textcolor{green}{\checkmark} \\
RoboPoint~\cite{yuan2024robopoint} & \textcolor{green}{\checkmark} \\
RoboPoint\textsc{-FT}~\cite{yuan2024robopoint} & \textcolor{green}{\checkmark} \\
Molmo~\cite{deitke2024molmopixmoopenweights} & \textcolor{green}{\checkmark} \\
GPT-4o~\cite{openai2022chatgpt} & \textcolor{green}{\checkmark} \\
\bottomrule
\end{tabular}
\end{minipage}

\vspace{0.3cm}

\begin{minipage}{0.4\linewidth}
\centering
\includegraphics[width=\textwidth]{supp_robot/experiments/9.png}
\end{minipage}%
\hfill
\begin{minipage}{0.58\linewidth}
\begin{tabular}{p{0.85\linewidth}c}
\toprule
\multicolumn{2}{p{\linewidth}}{\textbf{Question:} place the object inside the smallest box} \\
\midrule
LLaVa-Next~\cite{liu2023improvedllava} & \textcolor{red}{\texttimes} \\
LLaVa-Next\textsc{-FT}~\cite{liu2023improvedllava} & \textcolor{green}{\checkmark} \\
RoboPoint~\cite{yuan2024robopoint} & \textcolor{green}{\checkmark} \\
RoboPoint\textsc{-FT}~\cite{yuan2024robopoint} & \textcolor{green}{\checkmark} \\
Molmo~\cite{deitke2024molmopixmoopenweights} & \textcolor{green}{\checkmark} \\
GPT-4o~\cite{openai2022chatgpt} & \textcolor{red}{\texttimes} \\
\bottomrule
\end{tabular}
\end{minipage}
\end{minipage}

\end{minipage}

\vspace{0.3cm}

\begin{minipage}{\linewidth}
\centering

\begin{minipage}{0.48\linewidth}
\centering
\begin{minipage}{0.4\linewidth}
\centering
\includegraphics[width=\textwidth]{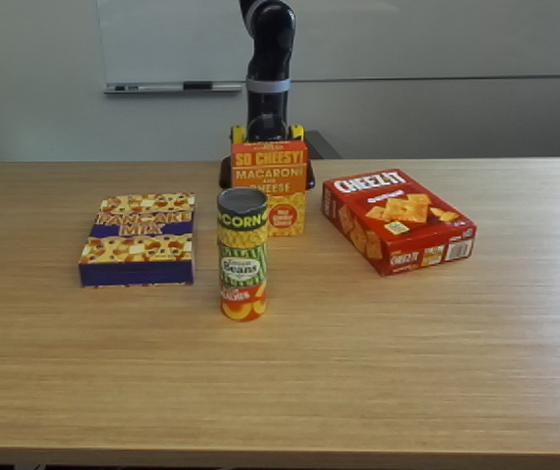}
\end{minipage}%
\hfill
\begin{minipage}{0.58\linewidth}
\centering
\begin{tabular}{p{0.85\linewidth}c}
\toprule
\multicolumn{2}{p{\linewidth}}{\textbf{Question:} can the robot directly pick the red orange peaches can without disturbing other objects?} \\
\midrule
LLaVa-Next~\cite{liu2023improvedllava} & \textcolor{green}{\checkmark} \\
LLaVa-Next\textsc{-FT}~\cite{liu2023improvedllava} & \textcolor{green}{\checkmark} \\
RoboPoint~\cite{yuan2024robopoint} & \textcolor{red}{\texttimes} \\
RoboPoint\textsc{-FT}~\cite{yuan2024robopoint} & \textcolor{red}{\texttimes} \\
Molmo~\cite{deitke2024molmopixmoopenweights} & \textcolor{green}{\checkmark} \\
GPT-4o~\cite{openai2022chatgpt} & \textcolor{green}{\checkmark} \\
\bottomrule
\end{tabular}
\end{minipage}

\vspace{0.3cm}

\begin{minipage}{0.4\linewidth}
\centering
\includegraphics[width=\textwidth]{supp_robot/experiments/7.png}
\end{minipage}%
\hfill
\begin{minipage}{0.58\linewidth}
\begin{tabular}{p{0.85\linewidth}c}
\toprule
\multicolumn{2}{p{\linewidth}}{\textbf{Question:} can the macaroni and cheese be placed on top of cheez-it without touching other objects?} \\
\midrule
LLaVa-Next~\cite{liu2023improvedllava} & \textcolor{red}{\texttimes} \\
LLaVa-Next\textsc{-FT}~\cite{liu2023improvedllava} & \textcolor{red}{\texttimes} \\
RoboPoint~\cite{yuan2024robopoint} & \textcolor{green}{\checkmark} \\
RoboPoint\textsc{-FT}~\cite{yuan2024robopoint} & \textcolor{green}{\checkmark} \\
Molmo~\cite{deitke2024molmopixmoopenweights} & \textcolor{red}{\texttimes} \\
GPT-4o~\cite{openai2022chatgpt} & \textcolor{green}{\checkmark} \\
\bottomrule
\end{tabular}
\end{minipage}

\vspace{0.3cm}

\begin{minipage}{0.4\linewidth}
\centering
\includegraphics[width=\textwidth]{supp_robot/experiments/7.png}
\end{minipage}%
\hfill
\begin{minipage}{0.58\linewidth}
\begin{tabular}{p{0.85\linewidth}c}
\toprule
\multicolumn{2}{p{\linewidth}}{\textbf{Question:} place on the object to the left of macaroni and cheese} \\
\midrule
LLaVa-Next~\cite{liu2023improvedllava} & \textcolor{red}{\texttimes} \\
LLaVa-Next\textsc{-FT}~\cite{liu2023improvedllava} & \textcolor{green}{\checkmark} \\
RoboPoint~\cite{yuan2024robopoint} & \textcolor{green}{\checkmark} \\
RoboPoint\textsc{-FT}~\cite{yuan2024robopoint} & \textcolor{green}{\checkmark} \\
Molmo~\cite{deitke2024molmopixmoopenweights} & \textcolor{green}{\checkmark} \\
GPT-4o~\cite{openai2022chatgpt} & \textcolor{red}{\texttimes} \\
\bottomrule
\end{tabular}
\end{minipage}
\end{minipage}%
\hfill
\begin{minipage}{0.48\linewidth}
\centering
\begin{minipage}{0.4\linewidth}
\centering
\includegraphics[width=\textwidth]{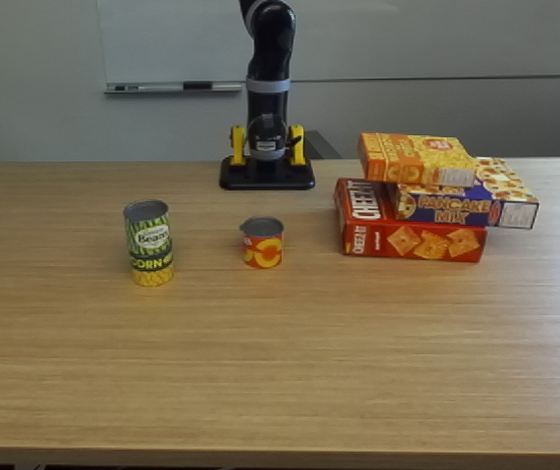}
\end{minipage}%
\hfill
\begin{minipage}{0.58\linewidth}
\centering
\begin{tabular}{p{0.85\linewidth}c}
\toprule
\multicolumn{2}{p{\linewidth}}{\textbf{Question:} is there an object that is not in a stack?} \\
\midrule
LLaVa-Next~\cite{liu2023improvedllava} & \textcolor{green}{\checkmark} \\
LLaVa-Next\textsc{-FT}~\cite{liu2023improvedllava} & \textcolor{green}{\checkmark} \\
RoboPoint~\cite{yuan2024robopoint} & \textcolor{green}{\checkmark} \\
RoboPoint\textsc{-FT}~\cite{yuan2024robopoint} & \textcolor{green}{\checkmark} \\
Molmo~\cite{deitke2024molmopixmoopenweights} & \textcolor{green}{\checkmark} \\
GPT-4o~\cite{openai2022chatgpt} & \textcolor{green}{\checkmark} \\
\bottomrule
\end{tabular}
\end{minipage}

\vspace{0.3cm}

\begin{minipage}{0.4\linewidth}
\centering
\includegraphics[width=\textwidth]{supp_robot/experiments/6.png}
\end{minipage}%
\hfill
\begin{minipage}{0.58\linewidth}
\begin{tabular}{p{0.85\linewidth}c}
\toprule
\multicolumn{2}{p{\linewidth}}{\textbf{Question:} is there space to place one of the cans on the cheez-it box?} \\
\midrule
LLaVa-Next~\cite{liu2023improvedllava} & \textcolor{red}{\texttimes} \\
LLaVa-Next\textsc{-FT}~\cite{liu2023improvedllava} & \textcolor{red}{\texttimes} \\
RoboPoint~\cite{yuan2024robopoint} & \textcolor{red}{\texttimes} \\
RoboPoint\textsc{-FT}~\cite{yuan2024robopoint} & \textcolor{red}{\texttimes} \\
Molmo~\cite{deitke2024molmopixmoopenweights} & \textcolor{red}{\texttimes} \\
GPT-4o~\cite{openai2022chatgpt} & \textcolor{red}{\texttimes} \\
\bottomrule
\end{tabular}
\end{minipage}

\vspace{0.3cm}

\begin{minipage}{0.4\linewidth}
\centering
\includegraphics[width=\textwidth]{supp_robot/experiments/6.png}
\end{minipage}%
\hfill
\begin{minipage}{0.58\linewidth}
\begin{tabular}{p{0.85\linewidth}c}
\toprule
\multicolumn{2}{p{\linewidth}}{\textbf{Question:} pick the highest object on the stack of two objects} \\
\midrule
LLaVa-Next~\cite{liu2023improvedllava} & \textcolor{red}{\texttimes} \\
LLaVa-Next\textsc{-FT}~\cite{liu2023improvedllava} & \textcolor{red}{\texttimes} \\
RoboPoint~\cite{yuan2024robopoint} & \textcolor{red}{\texttimes} \\
RoboPoint\textsc{-FT}~\cite{yuan2024robopoint} & \textcolor{red}{\texttimes} \\
Molmo~\cite{deitke2024molmopixmoopenweights} & \textcolor{red}{\texttimes} \\
GPT-4o~\cite{openai2022chatgpt} & \textcolor{red}{\texttimes} \\
\bottomrule
\end{tabular}
\end{minipage}
\end{minipage}

\end{minipage}

\end{scriptsize}

\caption{Additional robot experiments. A green check mark indicates that the model answered correctly. The \textsc{-FT} suffix denotes a model trained with \datasetname{}. The questions are purposely not cleaned to reflect realistic language inputs.}
\label{fig:supp_robot}
\end{figure*}

We present additional results from the robot experiments in~\cref{fig:supp_robot}. 
We observe that models trained with \datasetname{} consistently outperform baseline models in most cases, even though the prompt is not optimized for \datasetname{}-trained models. 
This demonstrates that the power of VLMs enables templated language to generalize to language unseen during training while maintaining spatial understanding capabilities. 
However, even with \datasetname{} training, the models struggle with understanding stacked items, indicating a need for further data augmentation with diverse layouts.
In a few cases, \datasetname{} training adversely affects performance, especially with RoboPoint~\cite{yuan2024robopoint}. 
We hypothesize that mixing the dataset with RoboPoint training data and \datasetname{} training data may lead to unforeseen side effects, particularly in grounding objects. 
Nevertheless, we demonstrate that \datasetname{} training enhances VLM's spatial understanding in real-life robotics experiments, even with freeform language.

\subsection{More Qualitative Examples}
\label{supp:qual}

\cref{fig:supp_qual} present additional qualitative comparisons between models trained on \datasetname{}.
The findings demonstrate that models trained on \datasetname{} consistently exhibit spatial understanding in the challenging \datasetname{}-Home dataset, even outperforming closed models like GPT-4o~\cite{openai2022chatgpt}.
However, we observed that object grounding is a crucial prerequisite for spatial understanding; the improvement is often hindered by the model's inability to ground objects in cluttered scenes, where GPT-4o performs more effectively.
Additionally, we show that the \datasetname{}-trained model successfully generalizes to unseen spatial relationships in BLINK-Spatial~\cite{fu2024blink}, including those involving distance, such as "touching." 

\newcommand{\coloreddot}[1]{%
    \begin{tikzpicture}[baseline=(C.base)]
        \node (C) at (0,0) {};
        \fill[#1] (0,0) circle (2.5pt);
        \draw[black, line width=1.5pt] (0,0) circle (1pt);
    \end{tikzpicture}%
}

\begin{figure*}[ht]
\centering
\captionsetup{width=0.9\textwidth} 
\begin{scriptsize}

\begin{minipage}{\linewidth}
\centering

\begin{minipage}{0.48\linewidth}
\centering
\begin{minipage}{0.4\linewidth}
\centering
\includegraphics[width=\textwidth]{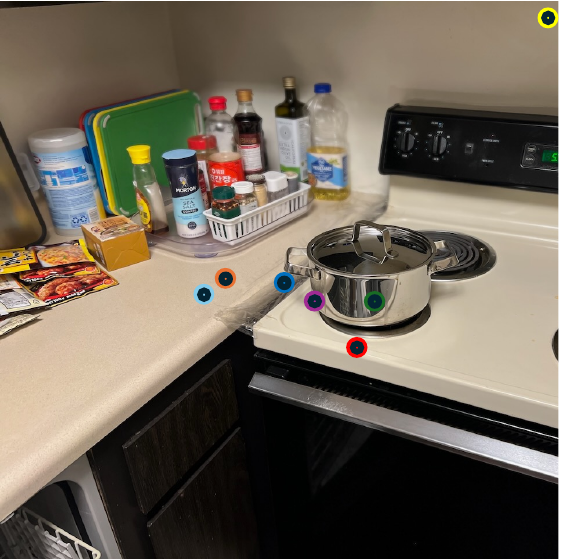}
\end{minipage}%
\hfill
\begin{minipage}{0.58\linewidth}
\centering
\begin{tabular}{p{0.85\linewidth}c}
\toprule
\multicolumn{2}{p{\linewidth}}{\textbf{Question:} Pinpoint several points within the vacant space situated to the left of the pot.} \\
\midrule
Answer & \coloreddot{Orange} \\
LLaVa-Next~\cite{liu2023improvedllava} & \coloreddot{Yellow} \\
LLaVa-Next\textsc{-FT}~\cite{liu2023improvedllava} & \coloreddot{CornflowerBlue} \\
RoboPoint~\cite{yuan2024robopoint} & \coloreddot{Purple} \\
RoboPoint\textsc{-FT}~\cite{yuan2024robopoint} & \coloreddot{RoyalBlue} \\
Molmo~\cite{deitke2024molmopixmoopenweights} & \coloreddot{red} \\
GPT-4o~\cite{openai2022chatgpt} & \coloreddot{Green}  \\
\bottomrule
\end{tabular}
\end{minipage}
\end{minipage}%
\hfill
\begin{minipage}{0.48\linewidth}
\centering
\begin{minipage}{0.4\linewidth}
\centering
\includegraphics[width=\textwidth]{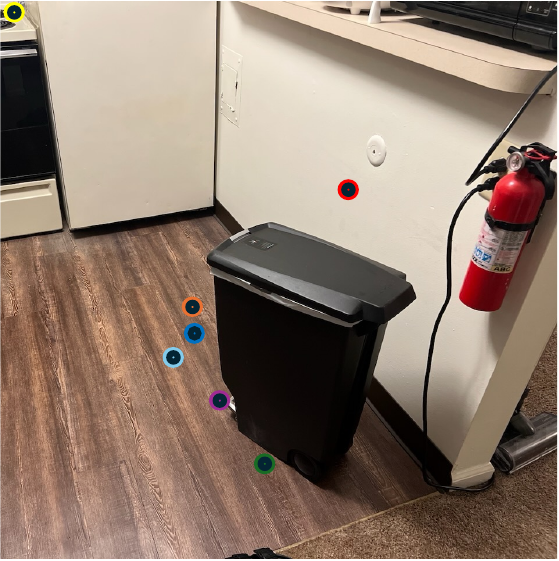}
\end{minipage}%
\hfill
\begin{minipage}{0.58\linewidth}
\centering
\begin{tabular}{p{0.85\linewidth}c}
\toprule
\multicolumn{2}{p{\linewidth}}{\textbf{Question:} Pinpoint several points within the vacant space situated behind the trash bin.} \\
\midrule
Answer & \coloreddot{Orange} \\
LLaVa-Next~\cite{liu2023improvedllava} & \coloreddot{Yellow} \\
LLaVa-Next\textsc{-FT}~\cite{liu2023improvedllava} & \coloreddot{CornflowerBlue} \\
RoboPoint~\cite{yuan2024robopoint} & \coloreddot{Purple} \\
RoboPoint\textsc{-FT}~\cite{yuan2024robopoint} & \coloreddot{RoyalBlue} \\
Molmo~\cite{deitke2024molmopixmoopenweights} & \coloreddot{red} \\
GPT-4o~\cite{openai2022chatgpt} & \coloreddot{Green}  \\
\bottomrule
\end{tabular}
\end{minipage}
\end{minipage}

\end{minipage}

\vspace{0.3cm}

\begin{minipage}{\linewidth}
\centering

\begin{minipage}{0.48\linewidth}
\centering
\begin{minipage}{0.4\linewidth}
\centering
\includegraphics[width=\textwidth]{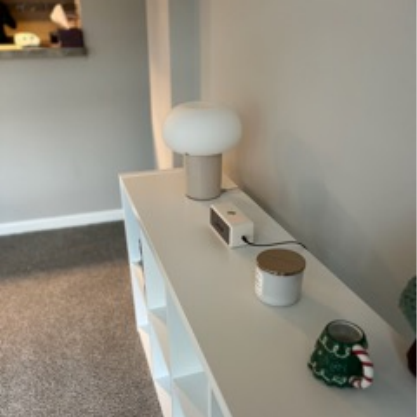}

\end{minipage}%
\hfill
\begin{minipage}{0.58\linewidth}
\centering
\begin{tabular}{p{0.85\linewidth}c}
\toprule
\multicolumn{2}{p{\linewidth}}{\textbf{Question:} Can the lamp fit in front of the shelf?} \\
\midrule
Answer & Yes \\
LLaVa-Next~\cite{liu2023improvedllava} & \textcolor{red}{\texttimes} \\
LLaVa-Next\textsc{-FT}~\cite{liu2023improvedllava} & \textcolor{green}{\checkmark} \\
RoboPoint~\cite{yuan2024robopoint} & \textcolor{red}{\texttimes} \\
RoboPoint\textsc{-FT}~\cite{yuan2024robopoint} & \textcolor{green}{\checkmark} \\
Molmo~\cite{deitke2024molmopixmoopenweights} & \textcolor{red}{\texttimes} \\
GPT-4o~\cite{openai2022chatgpt} & \textcolor{red}{\texttimes} \\
\bottomrule
\end{tabular}
\end{minipage}

\vspace{0.3cm}

\begin{minipage}{0.4\linewidth}
\centering
\includegraphics[width=\textwidth]{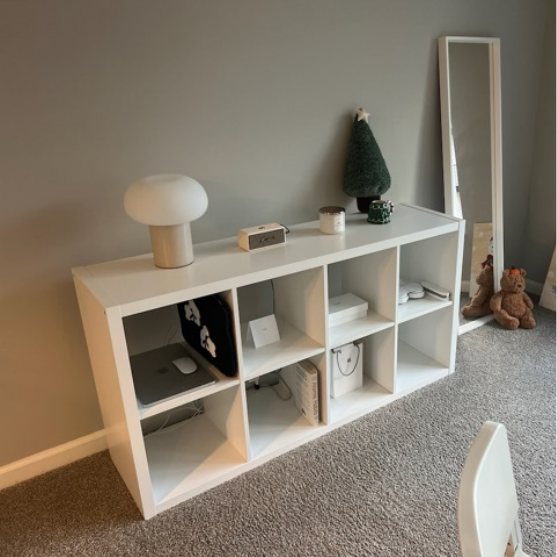}

\end{minipage}%
\hfill
\begin{minipage}{0.58\linewidth}
\begin{tabular}{p{0.85\linewidth}c}
\toprule
\multicolumn{2}{p{\linewidth}}{\textbf{Question:} Is the lamp above the shelf?} \\
\midrule
Answer & Yes \\
LLaVa-Next~\cite{liu2023improvedllava} & \textcolor{red}{\texttimes} \\
LLaVa-Next\textsc{-FT}~\cite{liu2023improvedllava} & \textcolor{green}{\checkmark} \\
RoboPoint~\cite{yuan2024robopoint} & \textcolor{red}{\texttimes} \\
RoboPoint\textsc{-FT}~\cite{yuan2024robopoint} & \textcolor{green}{\checkmark} \\
Molmo~\cite{deitke2024molmopixmoopenweights} & \textcolor{red}{\texttimes} \\
GPT-4o~\cite{openai2022chatgpt} & \textcolor{green}{\checkmark} \\
\bottomrule
\end{tabular}
\end{minipage}

\vspace{0.3cm}

\begin{minipage}{0.4\linewidth}
\centering
\includegraphics[width=\textwidth]{supp_qual/Picture7.pdf}
\end{minipage}%
\hfill
\begin{minipage}{0.58\linewidth}
\begin{tabular}{p{0.85\linewidth}c}
\toprule
\multicolumn{2}{p{\linewidth}}{\textbf{Question:} Is the dining table touching the donut?} \\
\midrule
Answer & Yes \\
LLaVa-Next~\cite{liu2023improvedllava} & \textcolor{red}{\texttimes} \\
LLaVa-Next\textsc{-FT}~\cite{liu2023improvedllava} & \textcolor{green}{\checkmark} \\
RoboPoint~\cite{yuan2024robopoint} & \textcolor{red}{\texttimes} \\
RoboPoint\textsc{-FT}~\cite{yuan2024robopoint} & \textcolor{green}{\checkmark} \\
Molmo~\cite{deitke2024molmopixmoopenweights} & \textcolor{red}{\texttimes} \\
GPT-4o~\cite{openai2022chatgpt} & \textcolor{red}{\texttimes} \\
\bottomrule
\end{tabular}
\end{minipage}
\end{minipage}%
\hfill
\begin{minipage}{0.48\linewidth}
\centering
\begin{minipage}{0.4\linewidth}
\centering
\includegraphics[width=\textwidth]{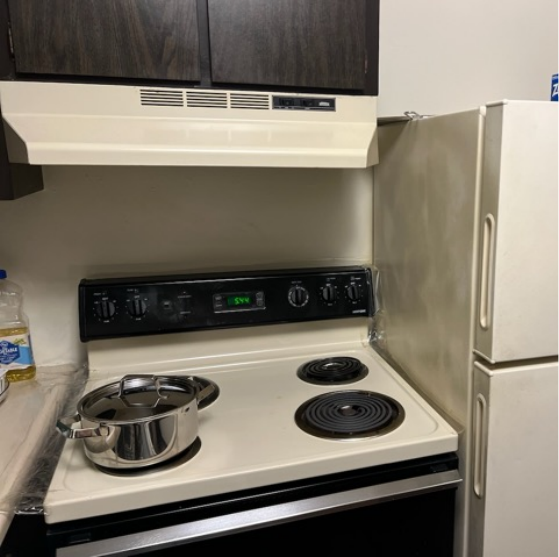}
\end{minipage}%
\hfill
\begin{minipage}{0.58\linewidth}
\centering
\begin{tabular}{p{0.85\linewidth}c}
\toprule
\multicolumn{2}{p{\linewidth}}{\textbf{Question:} Can the pot fit above the fridge?} \\
\midrule
Answer & Yes \\
LLaVa-Next~\cite{liu2023improvedllava} & \textcolor{red}{\texttimes} \\
LLaVa-Next\textsc{-FT}~\cite{liu2023improvedllava} & \textcolor{green}{\checkmark} \\
RoboPoint~\cite{yuan2024robopoint} & \textcolor{red}{\texttimes} \\
RoboPoint\textsc{-FT}~\cite{yuan2024robopoint} & \textcolor{green}{\checkmark} \\
Molmo~\cite{deitke2024molmopixmoopenweights} & \textcolor{red}{\texttimes} \\
GPT-4o~\cite{openai2022chatgpt} & \textcolor{red}{\texttimes} \\
\bottomrule
\end{tabular}
\end{minipage}

\vspace{0.3cm}

\begin{minipage}{0.4\linewidth}
\centering
\includegraphics[width=\textwidth]{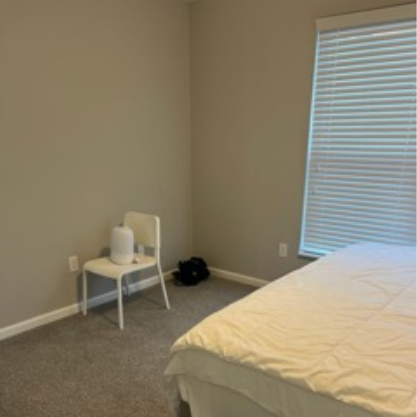}
\end{minipage}%
\hfill
\begin{minipage}{0.58\linewidth}
\begin{tabular}{p{0.85\linewidth}c}
\toprule
\multicolumn{2}{p{\linewidth}}{\textbf{Question:} Is the chair behind the bed?} \\
\midrule
Answer & Yes \\
LLaVa-Next~\cite{liu2023improvedllava} & \textcolor{red}{\texttimes} \\
LLaVa-Next\textsc{-FT}~\cite{liu2023improvedllava} & \textcolor{green}{\checkmark} \\
RoboPoint~\cite{yuan2024robopoint} & \textcolor{red}{\texttimes} \\
RoboPoint\textsc{-FT}~\cite{yuan2024robopoint} & \textcolor{green}{\checkmark} \\
Molmo~\cite{deitke2024molmopixmoopenweights} & \textcolor{red}{\texttimes} \\
GPT-4o~\cite{openai2022chatgpt} & \textcolor{red}{\texttimes} \\
\bottomrule
\end{tabular}
\end{minipage}

\vspace{0.3cm}

\begin{minipage}{0.4\linewidth}
\centering
\includegraphics[width=\textwidth]{supp_qual/Picture8.pdf}
\end{minipage}%
\hfill
\begin{minipage}{0.58\linewidth}
\begin{tabular}{p{0.85\linewidth}c}
\toprule
\multicolumn{2}{p{\linewidth}}{\textbf{Question:} Is the couch under the suitcase?} \\
\midrule
Answer & Yes \\
LLaVa-Next~\cite{liu2023improvedllava} & \textcolor{red}{\texttimes} \\
LLaVa-Next\textsc{-FT}~\cite{liu2023improvedllava} & \textcolor{green}{\checkmark} \\
RoboPoint~\cite{yuan2024robopoint} & \textcolor{red}{\texttimes} \\
RoboPoint\textsc{-FT}~\cite{yuan2024robopoint} & \textcolor{green}{\checkmark} \\
Molmo~\cite{deitke2024molmopixmoopenweights} & \textcolor{red}{\texttimes} \\
GPT-4o~\cite{openai2022chatgpt} & \textcolor{red}{\texttimes} \\
\bottomrule
\end{tabular}
\end{minipage}
\end{minipage}

\end{minipage}

\end{scriptsize}

\caption{Qualitative results on spatial reasoning benchmarks. The \textsc{-FT} suffix denotes a model trained with \datasetname{}. The first three rows show examples from \datasetname{}-Home, covering spatial context, spatial compatibility, and spatial configuration. For spatial context questions, only the first predicted point from each model is shown. The fourth row shows generalization to unseen spatial relationships on the Blink-Spatial~\cite{fu2024blink} dataset, demonstrating that the \datasetname{}-trained model can transfer to unseen relationships.}
\label{fig:supp_qual}
\end{figure*}

\end{document}